\documentclass[10pt,journal,compsoc, twoside]{IEEEtran}

\usepackage{cite}
\usepackage{amsmath}
\usepackage{algorithmic}
\usepackage{array}
\usepackage[caption=false,font=footnotesize]{subfig}
\usepackage{url}
\usepackage{fixltx2e}
\usepackage{graphicx}
\usepackage{tabularx, booktabs}
\usepackage{multirow}
\usepackage{amssymb}
\usepackage{commath}
\usepackage{float}
\usepackage{xcolor}

\newcommand{\change}[1]{\textcolor{black}{#1}}
\newcommand{\ignore}[1]{}
\newcolumntype{Y}{>{\centering\arraybackslash}X}
\newcolumntype{x}[1]{>{\centering\let\newline\\\arraybackslash\hspace{0pt}}p{#1}}

\providecommand{\ie}{\textit{i.e.}}

\providecommand{\etal}{\textit{et al.}}

\begin{document}

\title{\change{Deep Sketch-guided Cartoon Video Inbetweening}}

\author{Xiaoyu~Li,~Bo~Zhang,~Jing~Liao,~and~Pedro~V.~Sander
\IEEEcompsocitemizethanks{
\IEEEcompsocthanksitem X. Li is with the Department of Electronic and Computer Engineering, Hong Kong University of Science and Technology, Hong Kong.\protect\\
E-mail: xliea@connect.ust.hk.
\IEEEcompsocthanksitem B. Zhang is with Microsoft Research Asia, Beijing 100080, China.\protect\\ E-mail: Tony.Zhang@microsoft.com.
\IEEEcompsocthanksitem J. Liao is with the Department of Computer Science, City University of Hong Kong, Hong Kong. E-mail: jingliao@cityu.edu.hk.
\IEEEcompsocthanksitem P. V. Sander is with the Department of Computer Science and Engineering, Hong Kong University of Science and Technology, Hong Kong.\protect\\
E-mail: psander@cse.ust.hk.}
\thanks{Manuscript received April 19, 2005. \protect\\ (Corresponding author: Jing Liao.)}
}

\markboth{IEEE Transactions on Visualization and Computer Graphics,~Vol.~1, No.~1, August~2020}{Li \MakeLowercase{\textit{et al.}}: Deep Sketch-guided Cartoon Video Inbetweening}

\IEEEtitleabstractindextext{%
\begin{abstract}
We propose a novel framework to produce cartoon videos by fetching the color information from two input keyframes while following the animated motion guided by a user sketch. The key idea of the proposed approach is to estimate the dense cross-domain correspondence between the sketch and cartoon video frames, and employ a blending module with occlusion estimation to synthesize the middle frame guided by the sketch. After that, the input frames and the synthetic frame equipped with established correspondence are fed into an arbitrary-time frame interpolation pipeline to generate and refine additional inbetween frames. Finally, a module to preserve temporal consistency is employed. Compared to common frame interpolation methods, our approach can address frames with relatively large motion and also has the flexibility to enable users to control the generated video sequences by editing the sketch guidance. By explicitly considering the correspondence between frames and the sketch, we can achieve higher quality results than other image synthesis methods. Our results show that our system generalizes well to different movie frames, achieving better results than existing solutions.
\end{abstract}

}

\maketitle

\IEEEdisplaynontitleabstractindextext
\IEEEpeerreviewmaketitle

\IEEEraisesectionheading{\section{Introduction}\label{sec:introduction}}
%
%
%
%
\begin{figure*}
    \small
	\centering
	\includegraphics[width=\linewidth]{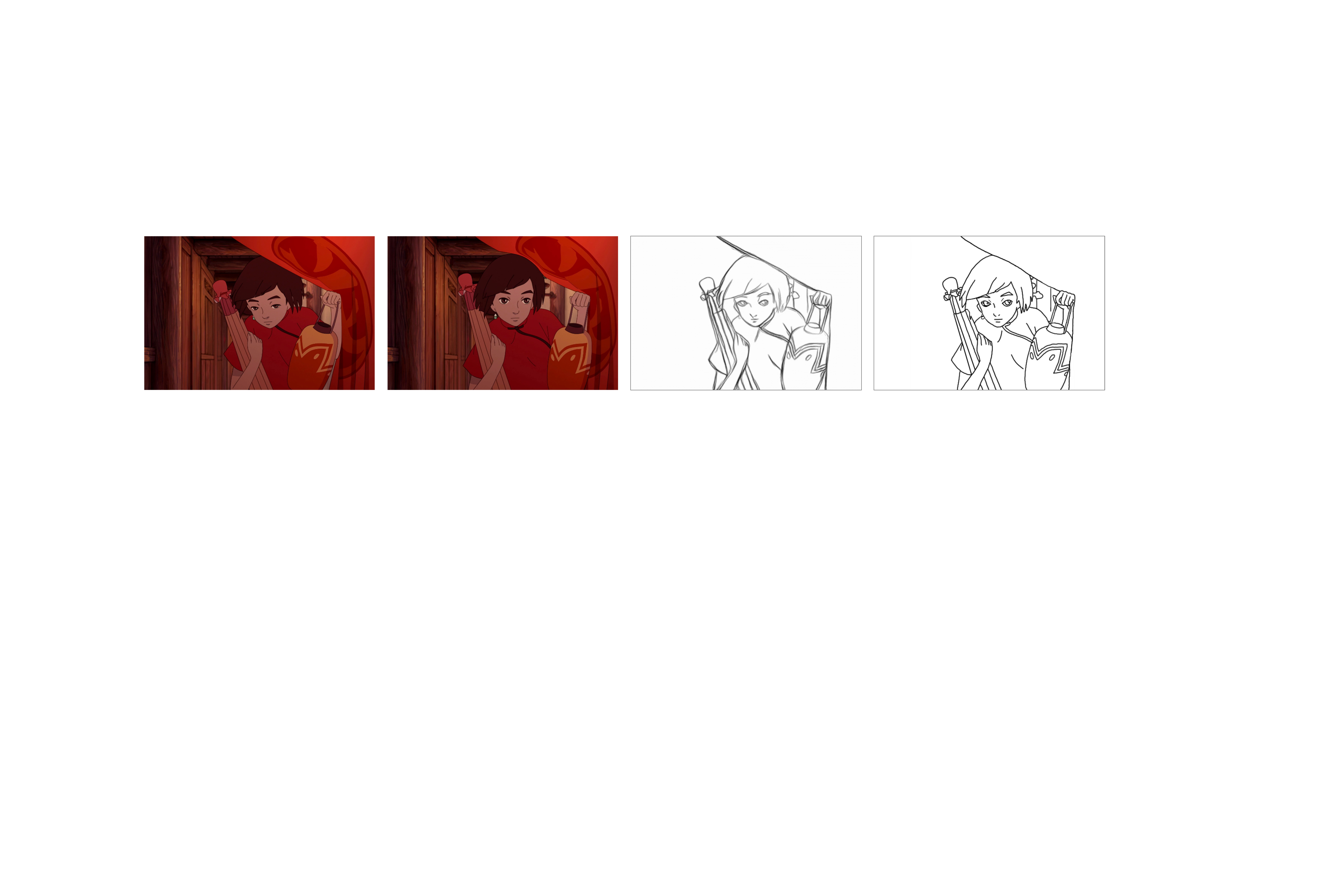}
	\begin{tabularx}{\linewidth}{YYYY}
      $I_0$ & $I_1$ & $S_{3/6}$ (rough) & $S_{3/6}$ (simplified)\\
    \end{tabularx}
	\includegraphics[width=\linewidth]{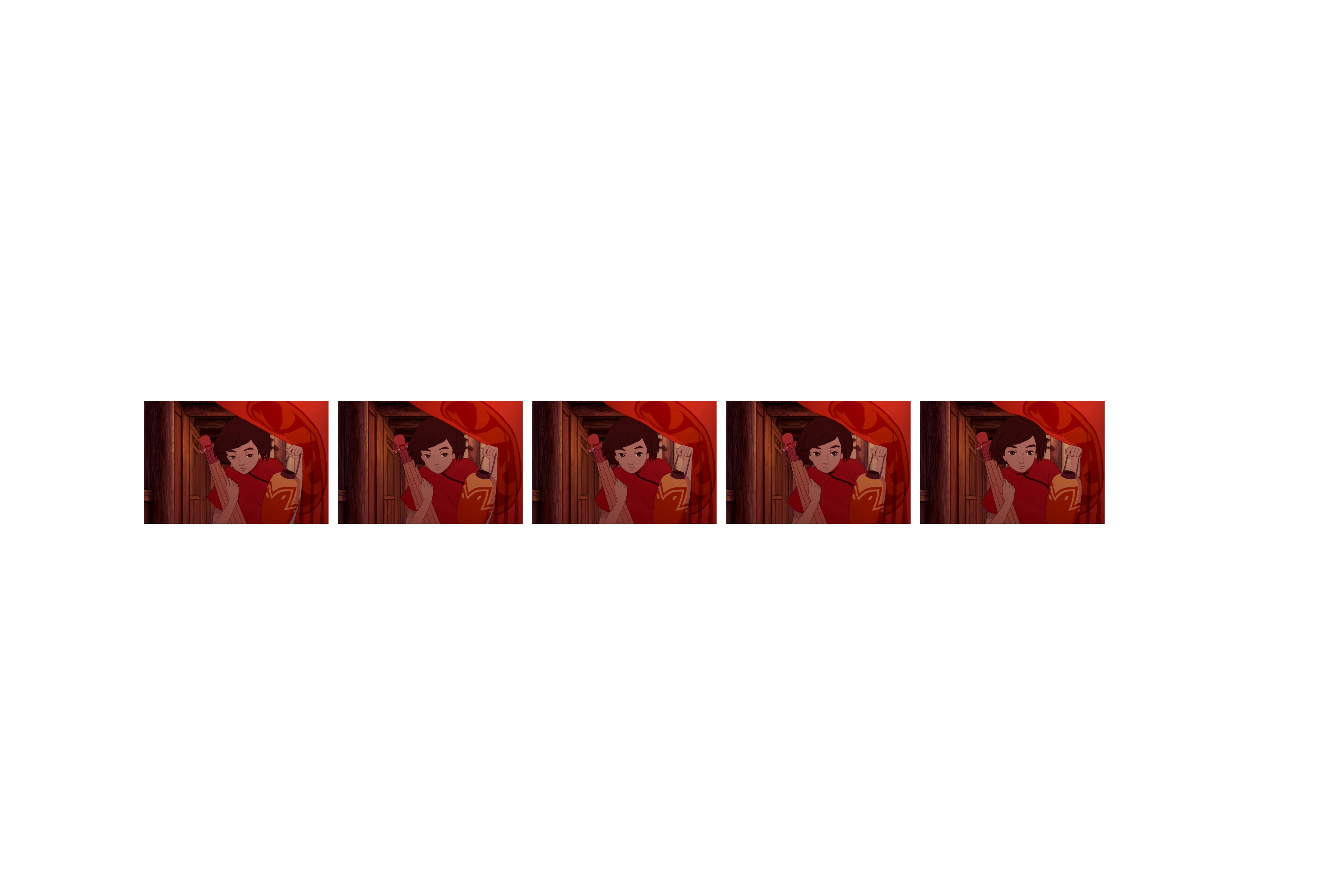}
	\begin{tabularx}{\linewidth}{YYYYY}
      $\hat I_{1/6}$ & $\hat I_{2/6}$ & $\hat I_{3/6}$ & $\hat I_{4/6}$ & $\hat I_{5/6}$\\
    \end{tabularx}
	\caption[width=\textwidth]{Our method synthesizes the frame $\hat I_{3/6}$ using two input keyframes $\{I_0, I_1\}$ and a guided sketch $S_{3/6}$ which was simplified from a rough input sketch drawn by artists. Furthermore, the approach can automatically interpolate additional inbetween frames \{$\hat I_{1/6}, \hat I_{2/6}, \hat I_{4/6}, \hat I_{5/6}$\} producing a smooth video with motion prescribed by the user-given sketch. \textcopyright B\&T.}
	\label{fig:teaserfigure}
\end{figure*}

\IEEEPARstart{C}{reating} cartoon animations can be partitioned into three main steps: drawing keyframes, inbetweening, and painting. First, an experienced animator draws the keyframes that capture the primary motion. Once completed, inbetweeners draw the inbetween frames for completing the motion, followed by a painter to fill the color in these sketches. Drawing and painting this large amount of inbetween frames is usually a specialized and time-consuming job, requiring intensive human labor from skilled professional artists, thus increasing the production cost. Therefore, we propose a system to alleviate this situation by automatically completing the inbetween frames including both the motion and color by only requiring some sketches for guidance, while maintaining the current animation workflow.

Research attempts have been made in helping users produce cartoon animations more easily. Sykora \etal~\cite{sykora2009lazybrush} propose an interactive tool which simplifies the sketch colorization process by filling the color within a region, but it still requires significant manual labor. Whited \etal~\cite{whited2010betweenit} present the BetweenIT system for the user-guided automation of tight inbetweening. Their methods mainly reduce the workload of drawing inbetween frames but not the painting process. We focus on completing the whole video considering both the motion and color. Some methods also utilize a hand-drawn sketch~\cite{zhu2016globally} or a color-coded skeleton~\cite{dvorovznak2018toonsynth} to guide synthetic animations, but Dvorožnák \etal~\cite{dvorovznak2018toonsynth} focus on one specific category of object and Zhu \etal~\cite{zhu2016globally} produce more free drawing animation but cannot handle motions with occlusions.

Furthermore, some related techniques can potentially be applied to assist the cartoon animation production, but many challenges restrict their direct use. One straightforward solution is to apply the state-of-the-art frame interpolation methods to two keyframes directly. However, these methods mainly focus on live-action (photorealistic) videos which makes it challenging to get satisfying results due to the large differences between live-action videos and cartoon animations~\cite{jiang2018super, bao2019depth}. More importantly, the artists hope to control the inbetweening by drawing rather than use the deterministic result from interpolation. Recent image synthesis methods, either for general purpose~\cite{isola2017image, wang2018high} or specifically for sketch colorization~\cite{zhang2018two, liu2018auto}, support automatically colorizing sketches with given frames as the reference. But without establishing the correspondence between the sketch and the frame, color bleeding artifacts may appear and the temporal consistency may also be hard to maintain.

The reason why it is hard to produce good results is that the problem of synthesizing videos from a sketch and cartoon keyframes is highly challenging. First, a cross-domain cartoon-to-sketch correspondence needs to be established. However, the cartoon frames are usually texture-less and the features in cartoon frames are unique and different from photographs, which is the target domain most previous matching methods are designed for. The situation is even worse for sketches, which makes establishing cartoon-to-sketch correspondence difficult. Second, the cartoon animations are more choppy and vigorous than live-action videos, which usually have unique object shape deformations and make occlusion estimation more difficult. Moreover, the unique contours in 2D cartoon frames can be easily destroyed by operations such as warping or resampling. It remains an open problem to generate cartoon frames without causing color bleeding or contour blurring. Finally, the valid frame rate (by removing the duplicated frames) of 2D cartoon animation is often low, i.e. 8-12 FPS, making it more challenging to achieve smooth interpolation results.

To address the above challenges and help users to automatically complete the inbetween frames, we propose a novel sketch-guided video synthesis system that can generate a sequence of inbetween frames controlled by one user-input sketch. Since the initial sketches from artists usually are very rough, casual, and potentially stylized, a pre-processing sketch cleanup needs to be performed to convert rough sketches into simplified clean line drawings. Then, the simplified sketches that contain the main contours or outlines of the objects are taken as the input guidance. To solve the cross-domain correspondence between a sketch and a cartoon frame, we first fill the large empty regions in the sketch with meaningful details by a transformation module conditioned on two keyframes. Then, two independent feature extractors are used to map the cartoon and sketch features into a common space that can be used to estimate the correspondence directly while maintaining the semantics of the original images. For occlusion handling, we estimate the occlusion mask by checking flow consistency and use a blending module to dynamically select and combine the pixels from two keyframes with these masks. Once the correspondence is established and the sketch frame is synthesized, an arbitrary-time frame interpolation module is used to generate and refine more inbetween frames. Finally, a temporal processing step is applied to further improve the result. Considering that the frame rate of 2D cartoon animation is low, we leverage the 3D cartoon movies which have smooth motions in nature to help training the interpolation and temporal processing module to produce temporally smooth 2D cartoon video results.

We demonstrate that our system can generate high-quality results in a broad range of scenes even containing some relatively large motions and works for cartoon movies with different styles. Moreover, with the sketch as guidance, our system allows the users to easily control the motion trajectory of the generated video by drawing sketches, thus increasing the flexibility. One example can be seen in Figure~\ref{fig:teaserfigure}. Our major contributions can be summarized as follows:

\begin{enumerate}
\item A system to resolve the demand in inbetweening by allowing users to specify the motion by drawing the sketch in a way that is compatible with the traditional animation workflow. We propose this new scenario and show that our method outperforms these existing possible solutions by a large margin.

\item A cross-domain correspondence estimation method for sketches and cartoon frames matching, achieving more accurate flow results than current optical flow estimation methods finetuned for this problem.

\item A blending method with a novel contour loss that better leverages the motion boundary clue to alleviate color bleeding, and an occlusion estimation module using flow consistency checking which is robust to the errors in estimated flows and occlusion masks.

\item An arbitrary-time frame interpolation pipeline and temporal processing module to produce and refine more inbetween frames with temporal coherence learned from 3D cartoon movies.
\end{enumerate}

\section{Related Work}
While there is no prior work that also tries to guide the synthesis of the whole 2D cartoon video using only one sketch, there are several techniques that can potentially be used to achieve this goal. In this section, we give an overview of those methods as well as the related works in cartoon animation.

\subsection{Sketch-guided Image Synthesis}
Sketches have been used to depict the visual world since prehistoric times and are deemed as a convenient art form to all humans~\cite{eitz2012humans}. Due to its simplicity, it can serve as a user-friendly control input for image synthesis. How to convert these easily acquired sketches to colorful images is thus a significant problem in both computational photography and cartoon animation. Chen \etal~\cite{chen2009sketch2photo} compose a realistic picture from a freehand sketch annotated with text labels, which is realized by stitching several text-related photographs discovered online. Eitz \etal~\cite{eitz2011photosketcher} and Bansal \etal~\cite{bansal2019shapes} adopt a similar approach which composite in-the-wild shapes and parts. However, methods in this category are not suitable to synthesize complex images due to the limited image database, and may often produce
disharmonious results as it is hard to unify the style of different parts. Recently, with the emergence of deep learning, sketches can be directly mapped to realistic photographs by learning from data~\cite{guccluturk2016convolutional, sangkloy2017scribbler, chen2018sketchygan}. Yet these works typically overfit a certain type of scene and usually produce low-resolution results with noticeable artifacts. Portenier \etal~\cite{portenier2018faceshop} present a sketch-guided image editing system that is specialized for faces. Moreover, recent image to image translation techniques can also be used to translate sketches to cartoon images~\cite{isola2017image, wang2018high, zhang2020cross}. However, applying these methods directly to our video task cannot give satisfactory result as the appearance of the output may deviate from the user-given keyframe and the generated video may introduce temporal flickering due to the nature of frame-by-frame processing. Furthermore, these methods are incapable to synthesize the frame at arbitrary intermediate time as in our approach.  

There are methods specifically designed for cartoon generation. Sykora \etal~\cite{sykora2009lazybrush} propose the first interactive tool that fills colors for sketch images. Zhang \etal~\cite{zhang2018two} and Liu \etal~\cite{liu2018auto} use deep neural networks to colorize sketches, but these methods target single images rather than video frames, and do not consider spatio-temporal consistency. The method proposed by Xing \etal~\cite{xing2015autocomplete} is similar to our scheme, which utilizes an artist-drawn sketch to animate a cartoon image. Nonetheless, they only consider 2D deformation to warp the input frame, and fail to address occlusions. We use two successive frames as input and can leverage richer information to address the issue. Dvorožnák \etal~\cite{dvorovznak2018toonsynth} also attempt to animate the cartoon frames, but their work is specialized to body skeletons, which limits their application to cartoon characters rather than the general genre. Instead of using a user-drawn image as guidance, Whited \etal~\cite{whited2010betweenit} propose to interpolate two keyframes by asking users to interactively match the outlines of input frames and manually adjust the motion trajectories. The method can achieve impressive results. However, the correspondence between frames has to be established manually, and the interpolated uniformly varying motion is not flexible enough. In comparison, our solution is more compatible to the cartoon inbetweening workflow and provides more freedom for artists to create desired motions. 

\subsection{Cross-domain Correspondence}
While significant advances have been made to estimate the optical flow for temporally adjacent frames~\cite{sun2018pwc, ilg2017flownet, dosovitskiy2015flownet, ranjan2017optical}, the semantic dense correspondence for general images remains challenging. Liu \etal~\cite{liu2010sift} and Yang \etal~\cite{yang2014daisy} rely on manually-crafted features to obtain the correspondence of scenes under large appearance variation. Ben-Zvi \etal~\cite{ben2016line} and Yang \etal~\cite{yang2018ftp} study the matching for stroke correspondence. Zhu \etal~\cite{zhu2016globally}, on the other hand, identify region correspondence between consecutive cartoon frames by solving a graph problem. However, this method assumes that the cartoon frame is composed of multiple flat regions with homogeneous color, and is thus not suitable to contemporary cartoon movies that usually contain complex shading and textures. There have been works that use deep neural networks for semantic correspondence. Liao \etal~\cite{liao2017visual} perform PatchMatch~\cite{barnes2009patchmatch} in a deep feature pyramid to compute the semantic dense correspondence. Aberman \etal~\cite{aberman2018neural}, on the other hand, focus on finding reliable sparse correspondence. However, both of them rely on a pre-trained classification model, e.g. VGG network, as feature extractor, and cannot capture semantics for sketch images. In our case, we wish to densely match the frame with the sketch image, where the latter lacks textures in most parts and only has semantic clues around the outlines. We solve this cross-domain correspondence problem in a self-supervised manner.

\change{Recently image translation methods provides the ability to translate the images across multiple domains by learning the domain-invariant representation from data~\cite{liu2017unsupervised, huang2018multimodal, liu2018unified, gonzalez2018image}. Liu \etal~\cite{liu2017unsupervised} map images in different domains to a latent code in a sheared-latent space. Huang \etal~\cite{huang2018multimodal} decomposes the image representation into a domain-invariant content code and a domain-specific style code. Liu\etal~\cite{liu2018unified} propose a model to learn disentangled features for describing cross-domain data to perform continuous cross-domain image translation and manipulation. All of these methods also inspire us to map the features of sketch image and cartoon frames to a domain-invariant space for correspondence estimation or alignment.}

\subsection{Video Frame Interpolation}
Video frame interpolation increases the video frame rate by inferring smooth motion and can be used for frame recovery in video streaming~\cite{huang2012cross, wu2015modeling} and slow motion effects~\cite{jiang2018super}. Classic frame interpolation algorithms are based on optical flow~\cite{barron1994performance, werlberger2011optical} and the quality of frame interpolation heavily depends on the flow accuracy. These methods usually require computationally expensive optimization and well-designed regularization~\cite{mahajan2009moving}. Recently, deep neural networks were proven to be a powerful hammer for frame interpolation and outperforms traditional methods in both quality and speed. Long \etal~\cite{long2016learning} first attempt to use deep neural network to directly synthesize the intermediate frames. Liu \etal~\cite{liu2017video} propose to learn a 3D optical flow in the space-time domain for frame warping and can support both frame interpolation and extrapolation. Many other learning strategies including interpolation kernels~\cite{niklaus2017video, niklaus2017videosep, bao2019memc}, context maps~\cite{niklaus2018context}, and incorporation of depth information~\cite{bao2019depth} can effectively improve the interpolation quality. Other methods focus on novel interpolation scenarios such as multi-frame interpolation for high frame rate videos~\cite{jiang2018super}, high resolution frame interpolation~\cite{peleg2019net} and the interpolation under camera shake~\cite{choi2019deep}. Yet all these methods can only produce deterministic results that appear plausible without any user control. In this work, we allow the user to explicitly control the motion path by drawing sketch, which we believe is the most convenient way for artist interaction. Since motion ambiguity is greatly reduced, our method demonstrates superior quality especially when processing long interval keyframes.

\section{Sketch-guided Video Synthesis}
\label{sec:method}

\begin{figure*}
	\centering
	\includegraphics[width=\linewidth]{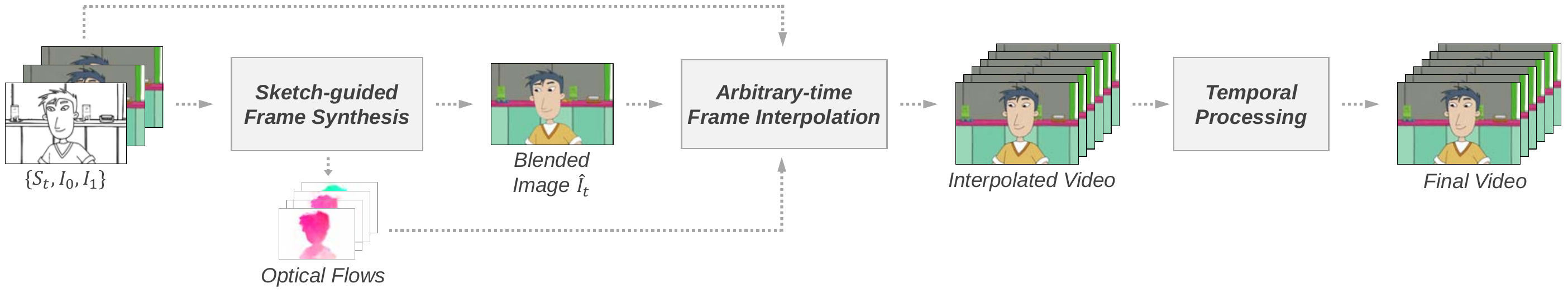}
	\caption{Our sketch-guided cartoon video synthesis consists of three stages. We first establish the correspondence between the sketch $S_t$ and keyframes $\{I_0,I_1\}$ and synthesize a blended image $\hat I_t$ corresponding to $S_t$. Then, we use the estimated flow from the first stage to interpolate additional inbetween frames and produce video results with an arbitrary frame rate. Finally, a temporal processing module is used to improve temporal consistency of the video.}
	\label{fig:over_sys}
\end{figure*}

We propose a sketch-guided cartoon video synthesis that utilizes one user-input sketch between a pair of keyframes to guide the motion in generated videos. Figure~\ref{fig:over_sys} shows an overview of our method. Given two consecutive cartoon keyframes $\{I_0, I_1\}\in \mathbb{R}^{H\times W \times 3}$ ($H$ and $W$ are image height and width respectively) and a sketch image $S_t\in \mathbb{R}^{H\times W}$ at the time $t \in (0,1)$, we first seek to synthesize an inbetween frame $\hat I_t$ that is geometrically aligned with the structure in $S_t$ and photometrically consistent with the input keyframes. Occlusions are also properly handled for $\hat{I}_t$ at this stage. Then, we use the estimated flow in the first stage to generate more inbetween frames at arbitrary intermediate times. Finally, a temporal processing network further reduces the artifacts by considering all the synthesized frames in spatio-temporal space, and finally produce smooth video results. We subsequently elaborate on each module.

\begin{figure}
	\centering
    \small
	\includegraphics[width=\linewidth]{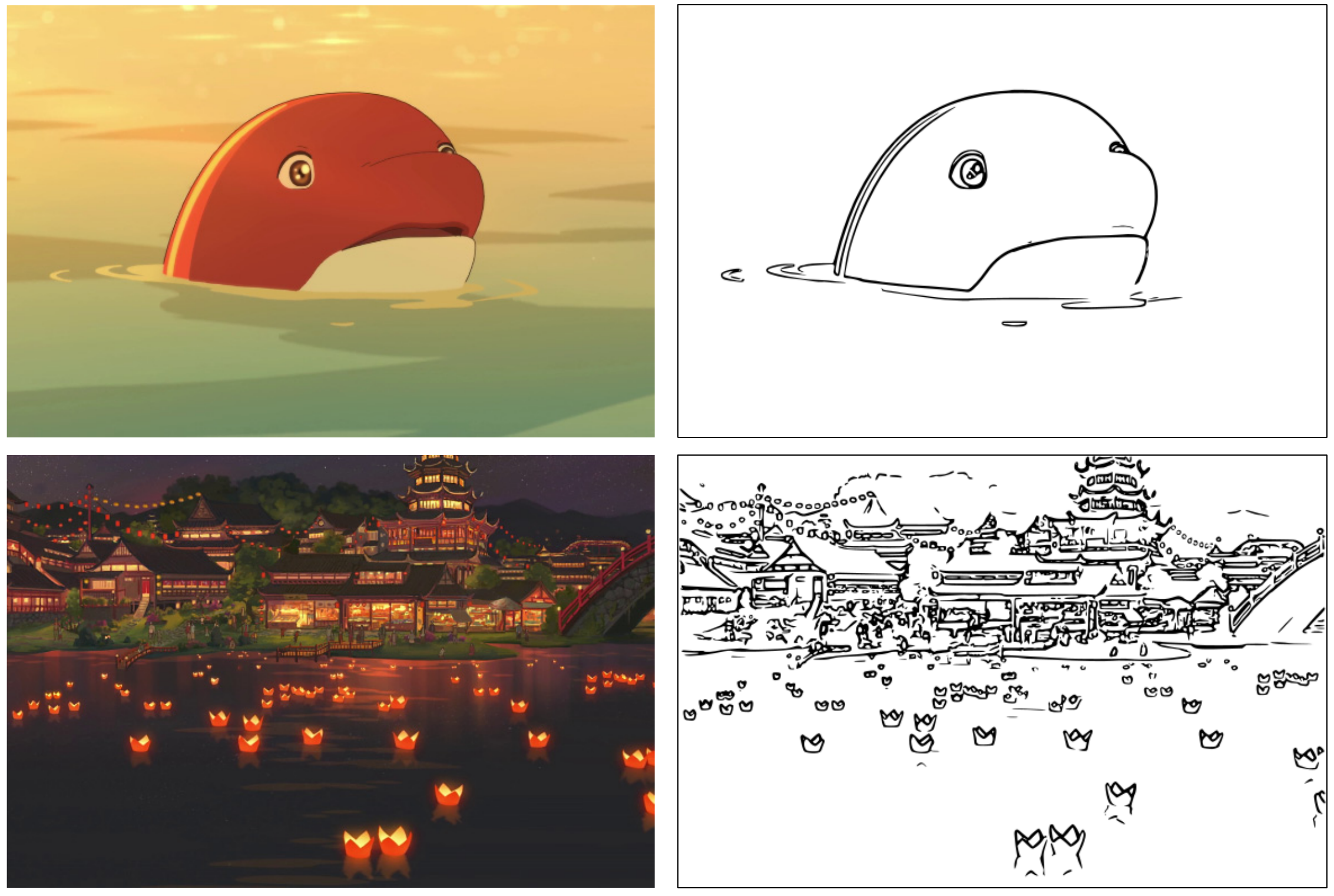}
	\begin{tabularx}{\linewidth}{YY}
        Video Frames & Synthetic Sketches \\
    \end{tabularx}
	\caption{Two synthetic sketch examples. The synthetic sketches outline the major content in the cartoon frame. \textcopyright B\&T.}
	\label{fig:sketches}
\end{figure}

\subsection{Sketch Simplification and Generation}
\label{sec:sketch}
Since the sketches drawn by artists can be rough and casual, developing a generic approach to process them directly is challenging. Therefore, a sketch simplification or cleanup is required as a pre-processing procedure, which removes superfluous details of sketches and leaves a clean line drawing to characterize the motion. In this work, we adopt existing simplification algorithms~\cite{simo2016learning, simo2018mastering} which are robust in producing good sketch simplification for unseen styles by leveraging unsupervised data during training. 
In our work, we use simplified sketches as the network input and focus more on video synthesis. We will use the term sketches to refer to the simplified ones which have a clean line drawing style unless otherwise specified.

In order to conduct supervised learning, we create a video dataset which contains cartoon frames $\{I_t\}$ and the corresponding synthetic sketch images $\{S_t\}$. It is well known that deep neural networks tend to overfit the
training data and may generalize poorly to images that slightly deviate from the training samples. Therefore, it is crucial to generate synthetic data as close as the hand-drawn sketches as possible. Our sketch generation procedure mostly follows Portenier \etal~\cite{portenier2018faceshop}. Specifically, we first extract contour maps using the holistically-nested contour detection (HED) method~\cite{xie2015holistically}, which provides multi-level contour map predictions. We choose the second level of its predictions as we empirically find that this level of output demonstrates good visual resemblance to real simplified sketches while maintaining high contour completeness. We further remove short contours by performing morphological operations. Additionally, we fit splines for the contour maps using Potrace~\cite{selinger6portrace} and smooth the curvature by manipulating control points as suggested in Portenier \etal~\cite{portenier2018faceshop}. Such curve smoothing is essential for improving the generalization since it allows better tolerance to the potentially inaccurate sketch simplification and helps the network to learn the synthesis based on rough contour locations. We show two examples of synthetic sketches in Figure~\ref{fig:sketches}. One can see that the synthetic sketches outline the major content in the cartoon frame and closely mimic the simplified sketch used during inference.

\subsection{Sketch-guided Frame Synthesis}
\label{sec:synthesis}

\begin{figure*}
	\centering
	\includegraphics[width=\linewidth]{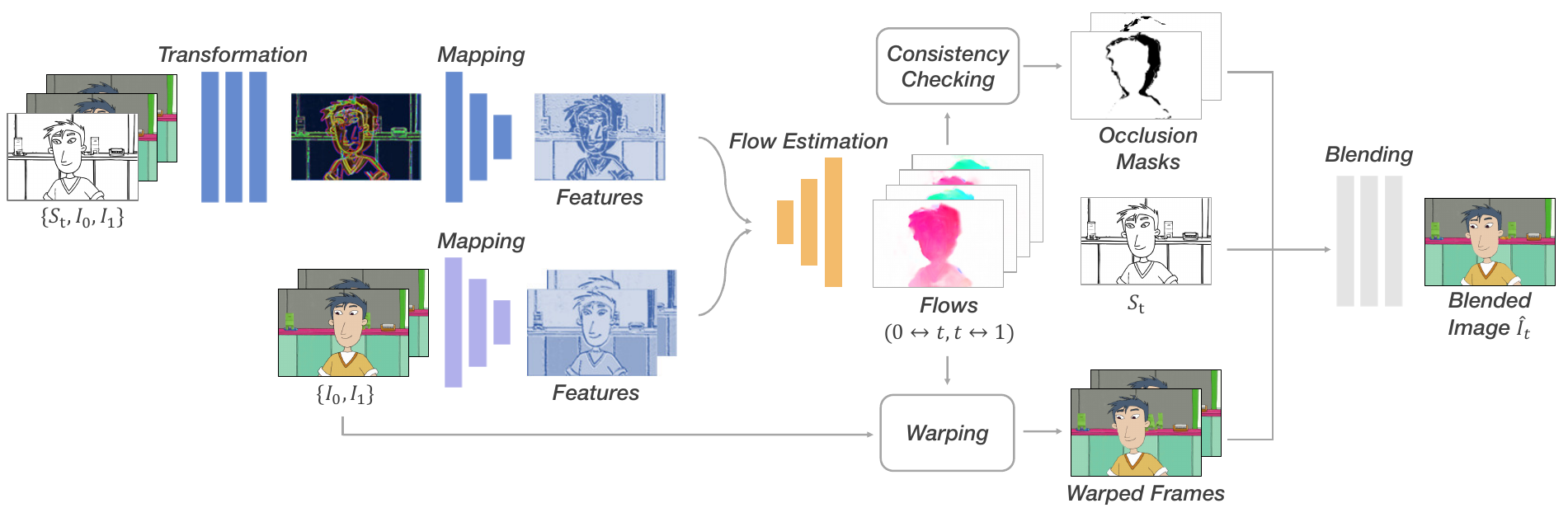}
	\caption{Overview of the sketch-guided frame synthesis pipeline, including cartoon-to-sketch correspondence estimation, occlusion handling by flow consistency checking and frame blending.}
	\label{fig:over_syn}
\end{figure*}

Given a simplified sketch $S_t$, we now aim to hallucinate the corresponding frame $\hat{I}_t$ which is also conditioned on the content images $\{I_0,I_1\}$. The framework of this sketch-guided frame synthesis is illustrated in Figure~\ref{fig:over_syn}. We first establish the dense correspondence between the sketch image and each of the input frames. Unlike conventional optical flow methods, we are trying to densely match images of distinct types. Then, we explicitly estimate a mask which accounts for the occluded region due to the foreground movement. This occlusion mask will guide the network to properly choose the non-occluded pixels from the warped frames and finally produce the blended result. Note that we learn these tasks in a self-supervised manner without any external labeling.
 
\subsubsection{Cartoon-to-sketch Correspondence}
Learning the cartoon-to-sketch correspondence is a non-trivial problem. As we will show in our experiments, directly using or fine-tuning an established flow estimation model fails to give accurate flow estimation since the sketch is a sparse representation and the correspondence for large areas of blank regions is essentially ill-posed. To accomplish reliable dense correspondence, both the sketch and the cartoon frame are expected to be mapped to a space where feature maps demonstrate detailed structures. To help the sketch to add these structures, we use cartoon frames as conditional inputs when extracting the features of sketch image. Specifically, we propose a transformer network with input $S_t$ and $\{I_0,I_1\}$ to hallucinate the missing structures of the sketch. After enhancing the structure of the sketch, we compute the correspondence in the deep features extracted from two independent mapping functions. We introduce this transformer only at the sketch branch, so the correspondence network has two asymmetric branches as shown in Figure~\ref{fig:over_syn}.

\paragraph{Architecture}
The transformer consists of several dilated residual layers~\cite{yu2017dilated} so that the receptive field is large enough to accommodate displacement between $S_t$ and $I_0$ (or $I_1$). After transforming the sketch image to a proper feature space, we adopt PWC-Net~\cite{sun2018pwc} as the flow estimator. This approach is capable of dealing with large motion by coarse-to-fine matching. Specifically, the PWC-Net estimates the flow in a feature pyramid, where the low-level flow is refined from a higher-level estimation. Here we initialize the feature extractor of PWC-Net with pre-trained weights but let the two branches independently update during training. The network computes the correlation in the cost volume~\cite{sun2018pwc} and estimates the bidirectional flow $f_{t \leftrightarrow 0}$ and $f_{t\leftrightarrow 1}$ for the two cartoon-sketch pairs. 

\paragraph{Loss Functions}
The ground truth flows are not available in our dataset, and flows computed by off-the-shelf flow estimation models may introduce errors. Instead, we use warping loss $\mathcal{L}_{warping}$ which calculates the $\ell_1$ difference between the ground truth and the warped frames according to the flow estimation. Albeit slight errors within occlusion, this loss suffices to serve as a rough guidance for flow training. The warping loss is defined as: 
\begin{equation}
\begin{aligned}
	\mathcal{L}_{warping} =\  &\norm{I_t - w(I_0, f_{t\to 0})}_1  +\  \norm{I_t - w(I_1, f_{t\to 1})}_1 \\
	& + \ \norm{I_0 - w(I_t, f_{0 \to t}) }_1 + \ \norm{I_1 - w(I_t, f_{1\to t}) }_1 
\end{aligned}
\end{equation}
where $w(\cdot,\cdot)$ denotes the backward warping function.

\subsubsection{Consistency Checking}

The foreground objects may undergo large displacements in two adjacent keyframes and inaccurate flows in the occlusion may severely degrade the warping quality. To alleviate this, we perform occlusion estimation by flow consistency checking. Occluded points cannot find corresponding counterparts in the other image, so the cyclic mapping will unlikely map them back to the original location. Formally, we use the spatial Euclidean distance to measure such consistency. Therefore the mask $O_{t \to 0}\in \mathbb{R}^{H\times W}$ accounting for the visibility in $I_0$ can be computed as:
\begin{equation}
O_{t \to 0}(p) = 2\sigma\left(||v(v(p, f_{t \to 0}), f_{0 \to t}) - p||_2\right) - 1
\label{eq:occlusion_mask}
\end{equation}
where $\sigma$ denotes the sigmoid function which is used to map the value in occlusion mask to $(0,1)$, and $v$ is the mapping function: $v(p, f) = p + f(p)$. The visibility of $I_1$ is computed similarly. Since the mask calculation is differentiable, it will in turn improve the flow prediction in the subsequent blending network.

\begin{figure*}
	\centering
	\includegraphics[width=\linewidth]{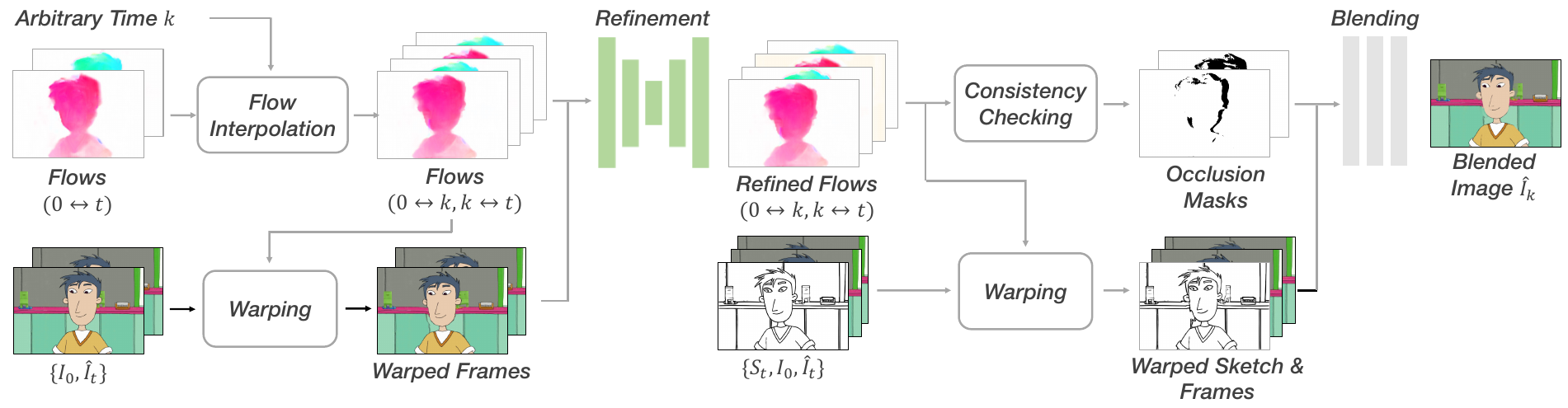}
	\caption{Overview of the arbitrary-time frame interpolation pipeline, including flow interpolation, flow refinement, occlusion handling by flow consistency checking and frame blending.}
	\label{fig:over_int}
\end{figure*}

\subsubsection{Blending}
We propose a blending network which predicts a soft blending mask $M \in \mathbb{R}^{H\times W}$ and fuses the warped cartoon frames $I_{t \to 0}$ and $I_{t \to 1}$ accordingly:
\begin{equation}
    \hat{I}_t = M \odot I_{t \to 0} + (1 - M) \odot I_{t \to 1}
\end{equation}
where $I_{t \to 0} = w(I_0, f_{t \to 0})$, $I_{t \to 1}=w(I_1, f_{t \to 1})$, and $\odot$ denotes the Hadamard product. The network takes as input the warped cartoon frames $\{I_{t\to 0},I_{t\to 1}\}$, the occlusion masks $\{O_{t \to 0},O_{t \to 1}\}$, and the sketch guidance $S_t$, and implicitly predicts the mask $M$ during the final blending. The blending mask should range in $[0,1]$ so it can be regarded as an attention map which properly selects from either frames. This blending mask not only considers the occlusion, but also resolves the blending artifacts due to the flow error. As each pixel in the blended image rigorously comes from the content frames, the output appears sharper than using a network that directly predicts a blended image.

\paragraph{Architecture} 
As the occlusion masks serve as a rough estimate, the network can predict the blending mask with a local receptive field. We determined that three convolutional layers are sufficient for a good estimation. 

\paragraph{Loss function}
The blending network needs to output $I_t$, so we introduce a blending loss to penalize the photometric $\ell_1$ error:
\begin{equation}
	\mathcal{L}_{blend} = \left\Vert \hat{I}_t-I_t \right\Vert_1
\end{equation}

The blended image, however, may still miss the contours that differentiate the neighboring color blocks, making the results appear blurry. This is because the contours are too thin to be penalized by the pixel-wise $\ell_1$ loss. In order to improve the perceptual sharpness and maintain the cartoon style, we propose to promote the contours by adopting a contour loss based on Chamfer matching. A similar loss function has previously been adopted for artist drawing synthesis~\cite{yi2019apdrawinggan}. The idea is to transform the target contour maps into distance maps through Euclidean distance transform, where each pixel value stores the distance to the closest contour, e.g., a larger value in the distance map means a further distance to the contours. Let $E(\hat{I}_t)\in\mathbb{R}^{H\times W}$ be the detected contour map of $\hat{I}_t$, and $D\in\mathbb{R}^{H\times W}$ be the distance map of the ground truth contours. In order to match contours to the ground truth, $(1-E(\hat{I}_t))$ should always sample small values in $D$. Formally, we penalize:
\begin{equation}
	\mathcal{L}_{contour} = \norm{(1-E(\hat{I}_t))\odot D}_1
\end{equation}
If $\hat{I}_t$ fails to produce contours at the expected location, it will induce a higher contour loss. In our implementation, we use HED~\cite{xie2015holistically} to detect contours for the outputs and the ground truth. Since the contour extraction $E$ is differentiable, the contour loss can guide the blending network to improve $\hat{I}_t$. 

So far, we have shown how to synthesize $\hat{I}_t$, by cartoon-sketch correspondence, occlusion handling by flow consistency checking, and frame blending as shown in Figure~\ref{fig:over_syn}. The overall objective function to train the entire synthesis network is:
\begin{equation}
\mathcal{L}_{syn} = \mathcal{L}_{blend} + \lambda_1 \mathcal{L}_{warping} + \lambda_2 \mathcal{L}_{contour}
\label{eq:allloss}
\end{equation}
where we empirically set $\lambda_1=0.5$ and $\lambda_2=0.01$.

\subsection{Arbitrary-time Frame Interpolation}

\label{sec:interpolation}
At this stage, the guided sketch frame has been synthesized and the correspondences have been established through the synthesis pipeline. Next, we will leverage this information in order to automatically interpolate more inbetween frames. We note that not all motions in cartoon animation can be interpolated this way due to the free drawing nature of cartoon frames. Thus, for more complex and larger motions, additional frames with guided sketches should be synthesized before performing interpolation. Nonetheless, frame interpolation in 2D cartoon video is very useful in many scenarios and simply using methods for live-action videos does not achieve satisfactory results. Instead we can directly leverage the already obtained flow information as well as some of the building blocks used for consistency checking and blending in the cartoon synthesis stage to produce a more accurate final result.

The interpolation pipeline for generating a frame at an arbitrary intermediate time is shown in Figure~\ref{fig:over_int}. Without loss of generality, we illustrate the interpolation of $I_k$ at time $k\in(0,t)$. We assume linear motion within $(0,t)$, so we approximate the bidirectional flow $f_{0 \leftrightarrow k}$ and $f_{k \leftrightarrow t}$ at time $k$ by scaling $f_{0 \leftrightarrow t}$ proportionally. These flows are then refined so as to suppress the motion artifacts near the object boundaries. Equipped with the estimated flow, the cartoon frame at that time can be synthesized with a procedure similar to that in Section~\ref{sec:synthesis}. 

\subsubsection{Flow Interpolation}
Since we assume linear motion from $I_0$ to $I_t$, for an arbitrary intermediate time $k$, the flow can be estimated by
\begin{equation}
	f_{0 \to k} = \frac{k}{t} f_{0 \to t},\quad 
	f_{t \to k} = \frac{t - k}{t} f_{t \to 0}
\end{equation}
Solving for the flows $f_{k \to 0}$ and $f_{k \to t}$ in opposite directions, however, is more problematic. Inspired by the work of Jiang \etal~\cite{jiang2018super}, we assume the optical flow is locally smooth. To compute the flow at time $k$, we can borrow the flow at the same position at time $0$ and $t$, and scale the magnitude proportionally. This way, we have the following two approximations:
\begin{equation}
    f_{k \to t}^0 \approx \frac{t - k}{t} f_{0 \to t}, \quad
	f_{k \to t}^1 \approx -\frac{t - k}{t} f_{t \to 0}
\end{equation}
Given these, we can combine them according to the temporal distance:
\begin{equation}
	f_{k \to t} = \frac{t-k}{t}f_{k \to t}^0 + \frac{k}{t}f_{k \to t}^1 \approx \frac{(t - k)^2}{t^2} f_{0 \to t} - \frac{k(t - k)}{t^2} f_{t \to 0} 
\end{equation}
Similarly, we derive the estimation of $f_{k \to 0}$ as
\begin{equation}
	f_{k \to 0} \approx -\frac{k(t - k)}{t^2} f_{0 \to t} + \frac{k^2}{t^2} f_{t \to 0} 
\end{equation}
One advantage of such interpolation scheme is that we can ensure the interpolated motion is temporally smooth.

\subsubsection{Flow Refinement}
The flow approximation works well for smooth motion, but may fail when points undergo non-linear motion or lie near motion boundaries. Thus, we train a flow refinement network to improve the flow estimation as shown in Figure~\ref{fig:over_int}. Specifically, the network takes as input the warped frames $\{I_{0\to k},I_{t\to k}\}$ and the rough flow estimation $\{f_{0 \leftrightarrow k}, f_{k \leftrightarrow t}\}$, and learns the flow residual for correction. Such residual learning helps to accelerate convergence and improve the flow quality. This flow refinement network adopts a U-Net structure~\cite{ronneberger2015u} that has a broader receptive field for flow refinement and thus gives globally consistent prediction.

\subsubsection{Frame interpolation} As shown in Figure~\ref{fig:over_int}, frame interpolation also performs consistency checking and final blending. The blending network shares the same weights those used for sketch-guided synthesis (Section~\ref{sec:synthesis}). One should notice that we use the warped sketch $S_t$, \ie, $S_{t\to k}$. This sketch information also improves this interpolation stage, as it helps to produce sharp results with clearly defined contours.

\subsection{Temporal Processing}
\label{sec:temporalprocessing}
The inbetween outputs are generated independently frame-by-frame. In order to further improve temporal consistency, we propose to optimize them using a temporal processing network that considers the entire space-time volume. Specifically, we use Unet~\cite{ronneberger2015u} to digest the concatenation of all the inbetween frames $\{I_k\}$ and provide a video output with improved consistency. This processing network adopts deformable convolutions~\cite{dai2017deformable} since the convolutional filters are performed on displaced grid points with learnable offsets, which can compensate the spatial misalignment of input video frames. We consistently observed improved loss curves using deformable convolution for temporal processing. The network is optimized to reduce the $\ell_1$ loss between the ground truth and the generated video. Later we will see that this processing step also helps reduce the spatial artifacts, since the networks can rely on motion trajectory to propagate more information to regions with unreliable colors.

\section{Training}
\label{sec:training}

We next present how we construct the dataset and some strategies we adopt for training, both of which play an important role in our methods.

\subsection{Data Preparation}
We first extract all the frames in a 2D cartoon movie called Spirited Away, a 24fps HD animated film with a variety of different scenes. Since the movie usually has some repeated frames or frames with only subtle variations that have limited contribution to learning the model, we prune neighbor frames with an SSIM of 0.95 or higher. Finally, we produce a temporally downsampled movie with an average frame rate of approximately 8fps.

We then sort these frames by scenes. However, not all scenes are applicable for frame synthesis. Scenes without distinct semantic correspondences like rainfall and rising smoke are very challenging. Keeping these examples in the training data reduces the training performance. As such, we calculate matching costs for each pixel with its corresponding pixels at adjacent frame to filter the scenes. More specifically, we divide each scene into multiple triples of frames. In every triple, the first and third frames are warped to the second frame using optical flow PWC-Net~\cite{sun2018pwc}. We select the pixels which are closer to the second frame based on $\ell_1$ error to form the final warped second frame. We then calculate the number of pixels which have a $\ell_1$ error less than 5\% of the color range between the final warped frame and the ground truth frame. Finally, the scenes with a pixel matching rate of less than 65\% will be removed from the data. \change{These selective frames will be used to generate their corresponding sketch images using the method described in Section~\ref{sec:sketch}.}

For arbitrary-time frame interpolation pipeline and temporal processing, smaller smooth motion videos are required. Unfortunately, the common 2D cartoon animations can not meet these requirements. Because “one-shot three frames” or “one-shot two frames” are used for saving cost, most 2D animations are limited animated to 8-12 fps if the repetitive frames are removed. Therefore, it can not be used as the dataset to help our system to generate the video with an arbitrary high frame rate. To address this problem, we create a smooth motion dataset based on 3D cartoon videos. We explored with different options and used a 3D cartoon called The Octonauts, which is a 25fps full animated video.\ignore{without any repetitive frames.} Next, we introduce how we use these two datasets to train the system.

\subsection{Training Procedure}
The entire system can be trained end-to-end by optimizing the networks from scratch in a single stage. However, it is difficult to get a good result in practice due to a large number of components and intermediate results. Moreover, recent works~\cite{xia2018invertible, zhang2018two, bao2019depth} have shown that multi-stage training and pre-trained models can be beneficial in this scenario. Therefore, we adopt a two-stage training to optimize the full model, which is shown to be more effective in our ablation study. We first train the frame synthesis pipeline with the correspondence network and the blending network as the two tasks mutually benefit each other using our 2D cartoon dataset and loss function in Equation~\ref{eq:allloss}. Then, we jointly train all the modules using the 3D dataset and $\ell_1$ loss between generated video and the ground truth video. For the first stage, we use 3 consecutive frames as one sample to synthesize the middle frame. For the second stage, we use 7 consecutive frames as a sample to synthesize one middle frame and interpolate the remaining four middle frames. During inference, we can interpolate an arbitrary number of frames.

Our method is implemented in PyTorch~\cite{paszke2019pytorch} and the code will be made publicly available. We utilize approximately 36,000 2D cartoon frames and 10,000 3D cartoon frames for training. While all the frames come from one 2D cartoon movie and one 3D cartoon teleplay, we will show that it has the ability to generalize to many different movies and cartoon styles. We train our network on frames with resolution of $384 \times 576$ using the Adam optimizer~\cite{kingma2014adam} with $\beta_1 = 0.9 $ and $\beta_2 = 0.999$, a learning rate of 0.0001 without any decay schedule, and batch size of 4 samples. It takes approximately 60 epochs to converge for the first stage and 100 epochs for the jointly trained second stage. The entire training procedure takes approximately three days on 4x Tesla M40 GPUs.

\section{Results}

We next present ablation studies to verify the benefits of key components of our method, followed by comparisons to related techniques, including flow estimation, frame interpolation, and image synthesis methods. Finally, we present additional results to show the generalization ability of our method.


\subsection{Model Analysis}

We construct two test datasets for our ablation study. For {\em 2D Cartoon Clips}, we collect 30 cartoon clips from 2D cartoon animations which are not seen during training. These clips cover different styles and scenes. We also downsample these clips temporally and generate a sketch for each middle frame of each triple as we did for training data preparation. Finally, we get approximately 500 triples for testing in this dataset. For {\em 3D Cartoon Clips}, we use 20 cartoon clips from a different 3D TV animation to evaluate final generated videos with a smooth transition. We take every 7 
consecutive frames as a group and only generate one sketch in the middle frame for each group having a total of 120 groups. We use PSNR, SSIM, and $\ell_1$ between the estimated frames and the ground truth frames as the evaluation criteria. 

\begin{table}
\centering
\small
\caption{Ablation study for the first stage (frame synthesis pipeline).}
\vspace{-0.3cm}
\label{tab:tab_syn}
\setlength{\tabcolsep}{6pt}
\begin{tabular}{lccc}
	\toprule
	\multicolumn{1}{c}{\multirow{2}{*}{Model}} & \multicolumn{3}{c}{2D Cartoon Clips} \\
	\cmidrule(lr){2-4} 
    \multicolumn{1}{c}{} & $\ell_1$ Loss & PSNR & SSIM \\
	 
	 \cmidrule(lr){1-4}
	\change{w/o sketch image} & \change{0.0219} & \change{25.97} & \change{0.871} \\
	w/o transformer         & 0.0101 & 32.40 & 0.945 \\
	w/o occlusion mask      & 0.0104 & 32.02 & 0.944 \\
	w/o warping loss        & 0.0107 & 31.74 & 0.937 \\
	w/o contour loss        & 0.0102 & 32.19 & 0.945 \\
	\cmidrule(lr){1-4}
	full synthesis model &  \textbf{0.0095} & \textbf{32.70} & \textbf{0.950} \\
	\bottomrule
\end{tabular}
\end{table}

\begin{table}
\centering
\small
\caption{Ablation study for joint training the entire framework.}
\vspace{-0.3cm}
\label{tab:tab_int}
\setlength{\tabcolsep}{6pt}
\begin{tabular}{lccc}
	\toprule
	\multicolumn{1}{c}{\multirow{2}{*}{Model}}  & \multicolumn{3}{c}{3D Cartoon Clips}\\
	
	\cmidrule(lr){2-4}
	
    \multicolumn{1}{c}{} & $\ell_1$ Loss & PSNR & SSIM \\
	 
	 \cmidrule(lr){1-4}
	    w/o joint training  & 0.0113 & 29.90 & 0.949 \\
		w/o refinement  & 0.0106 & 30.04 & 0.951 \\
		w/o temporal processing & 0.0104 & \textbf{30.11} & 0.952 \\
	\cmidrule(lr){1-4}
	full model  & \textbf{0.0102} & \textbf{30.11} & \textbf{0.953} \\
	\bottomrule
\end{tabular}
\end{table}

\begin{figure*}
    \small
	\centering
	\includegraphics[width=\linewidth]{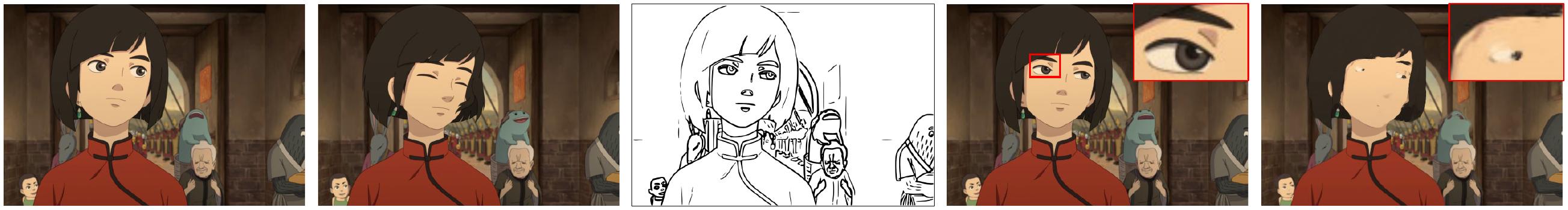}
	\begin{tabularx}{\linewidth}{YYYYY}
        $I_0$ & $I_1$ & $S_t$ & $I_t$ (ground truth) & \change{w/o sketch image}\\
    \end{tabularx}
	\includegraphics[width=\linewidth]{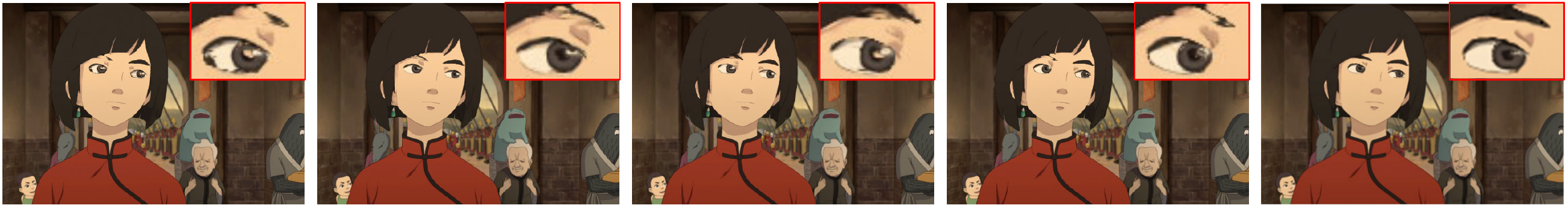}
	\begin{tabularx}{\linewidth}{YYYYY}
          w/o occlusion mask & w/o transformer & w/o warping loss  & w/o contour loss & synthesis model\\
    \end{tabularx}
	\caption{Sample frames for the ablation study of frame synthesis stage. Removing any of the components causes performance degradation of varying degrees. \textcopyright B\&T.}
	\label{fig:fig_ablation}
\end{figure*}

\begin{figure*}
    \small
	\centering
	\includegraphics[width=\linewidth]{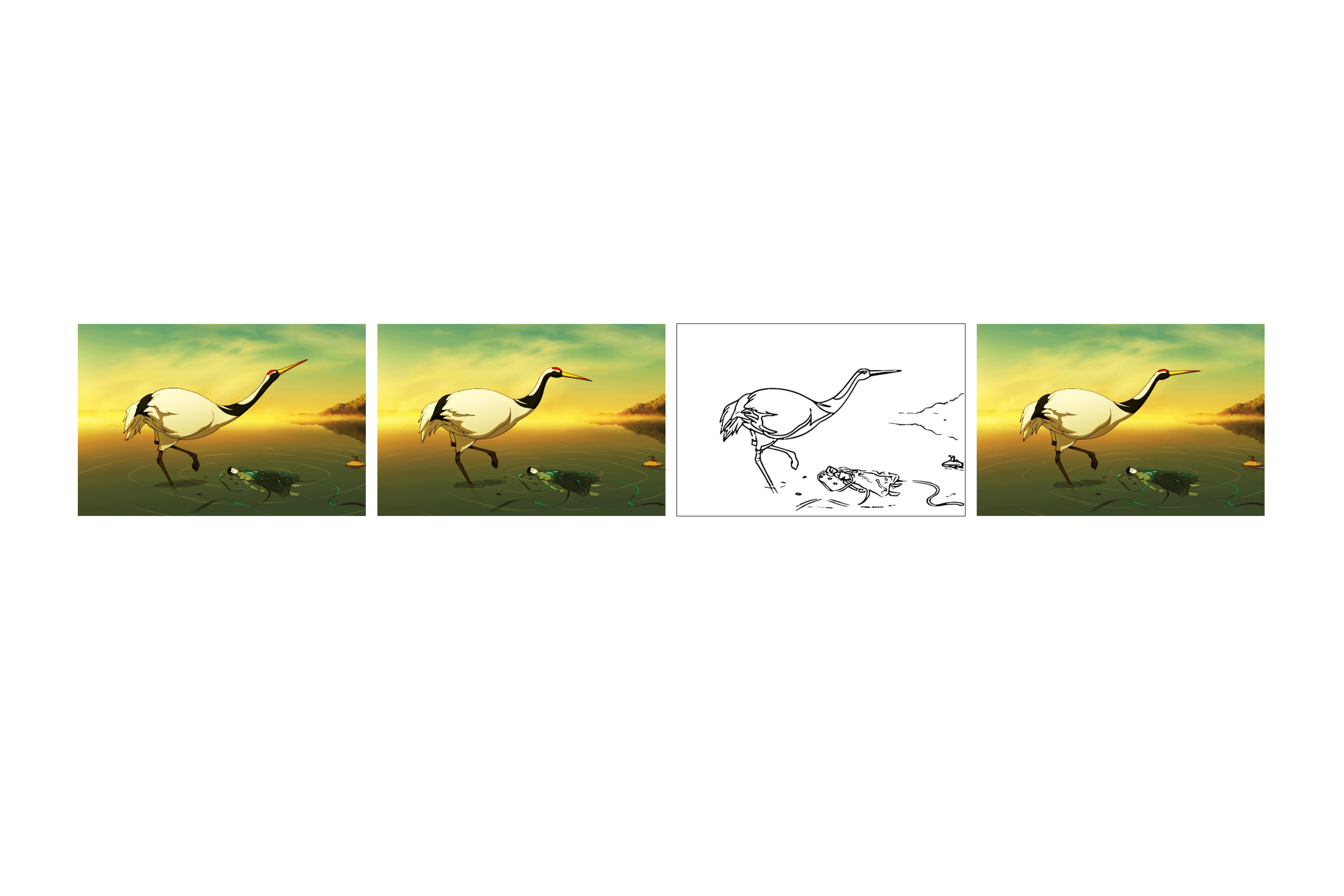}
	\begin{tabularx}{\linewidth}{YYYY}
        $I_0$ & $I_1$ & $S_{3/6}$ & $I_{3/6}$ (ground truth) \\
    \end{tabularx}
	\includegraphics[width=\linewidth]{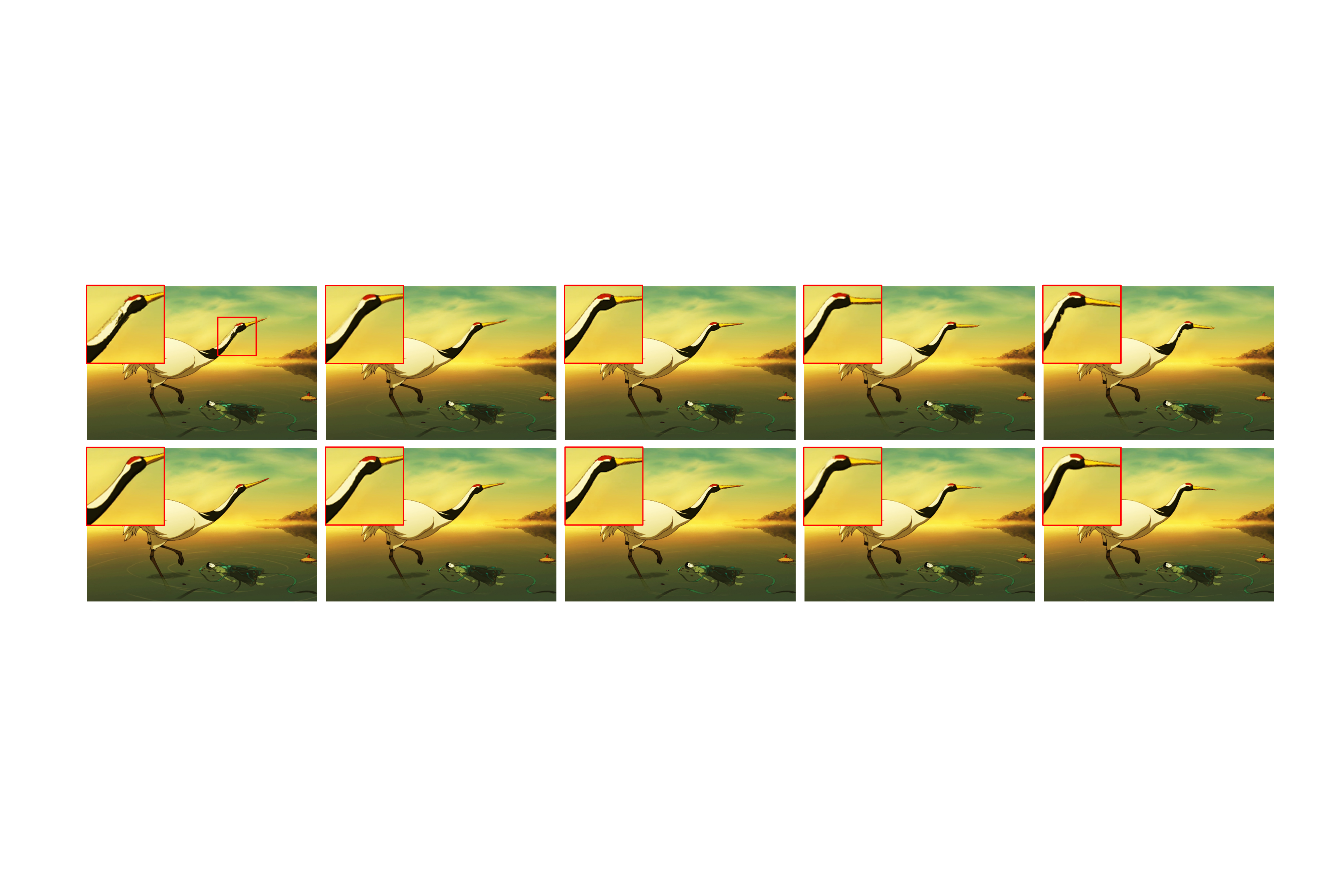}
	\begin{tabularx}{\linewidth}{YYYYY}
        $\hat I_{1/6}$ & $\hat I_{2/6}$ & $\hat I_{3/6}$ & $\hat I_{4/6}$ & $\hat I_{5/6}$\\
    \end{tabularx}
	\caption[width=\textwidth]{Example results comparing the model without refinement and temporal processing (second row) with our full model with joint training of the entire framework (third row). We can see that the full model can produce higher quality results equipped with the refinement and temporal processing. \textcopyright B\&T.}
	\label{fig:comp_refine}
\end{figure*}

We first conduct an ablation study to determine the optimal settings for our sketch-guided frame synthesis pipeline in the first stage of training. The second stage that jointly trains all the modules starts from the first stage results by utilizing its synthetic middle frame and established correspondences. Therefore, the key empirical observation is that the better results the first stage can achieve, the better performance we get from the joint training of the second stage. We verify the effectiveness of four components in the first stage: transformer, occlusion mask, warping loss and contour loss. \change{We also do an ablation study without the sketch image as the guidance. To keep all the other modules intact, we remove the sketch image by using a blank one}. As shown in Table~\ref{tab:tab_syn}, the best results are achieved in terms of $\ell_1$ loss, EPE and PSNR when the full synthesis model is used. Removing any of them causes performance degradation of varying degrees. An image example from the test dataset is shown in Figure~\ref{fig:fig_ablation}. In this example, our method can produce results that have fewer visible artifacts and a high-quality synthesis result. The small deviation from the ground truth frame $I_t$ is due to the guided sketch $S_t$ does not follow the edges of $I_t$ accurately. We find that warping loss and occlusion masks can increase the overall performance by a relatively large margin. The transformer can improve the region without dense guided strokes and contour loss can help maintain boundaries. \change{Without the sketch as the guidance, the network has difficulty learning the motion in the 2D cartoon animation correctly, resulting in misalignment with respect to the ground truth.}

Next, we evaluate the training strategies and key components for the second stage in which we jointly train all the modules after initialing the parameters trained from stage one. As the first experiment, we load the parameters of the first stage but do not update them and only train the interpolation network and temporal processing network (referred to as {\em w/o joint training}). Then we selectively remove the flow refinement and temporal processing stage to show their effect in results. Finally, we show the results using the full model. All of the results are summarized in Table~\ref{tab:tab_int}. Note that the full model outperforms all the other settings. An image example is shown in Figure~\ref{fig:comp_refine}. We can see that with the refinement and temporal processing, our method can produce higher quality and more temporally coherent results.


\subsection{Comparison to Flow Estimation Methods}
We first compare our cartoon-to-sketch correspondence method to recent advanced flow estimation methods PWC-Net~\cite{sun2018pwc}. Since the ground truth optical flows for 2D cartoon animations are unavailable, we use the warping loss to train and evaluate the flows in 2D cartoon clips. More specifically, we use the estimated flows to warp one frame to another and calculate the $\ell_1$ loss between warped frames and ground truth frames. We also use the MPI {\em Sintel} Flow Dataset~\cite{Butler2012} which is used for the evaluation of optical flow derived from the open-source 3D animated short film, Sintel. Notice that this dataset has a significantly different style from the training set. Since the clips from this dataset already have large motion, we do not temporally downsample the clips and only generate the sketch to provide it as input. The Sintel dataset contains 341 triples for our testing. Because the ground truth flow is available, we also use the endpoint error (EPE) as a metric. \change{For each pair of frames, we take a frame image and a sketch image from the other frame as inputs to estimate their optical flow. We use our sketch generation method in Section~\ref{sec:sketch} to generate the corresponding sketch images.}

\begin{figure}
    \small
	\centering
	\includegraphics[width=\linewidth]{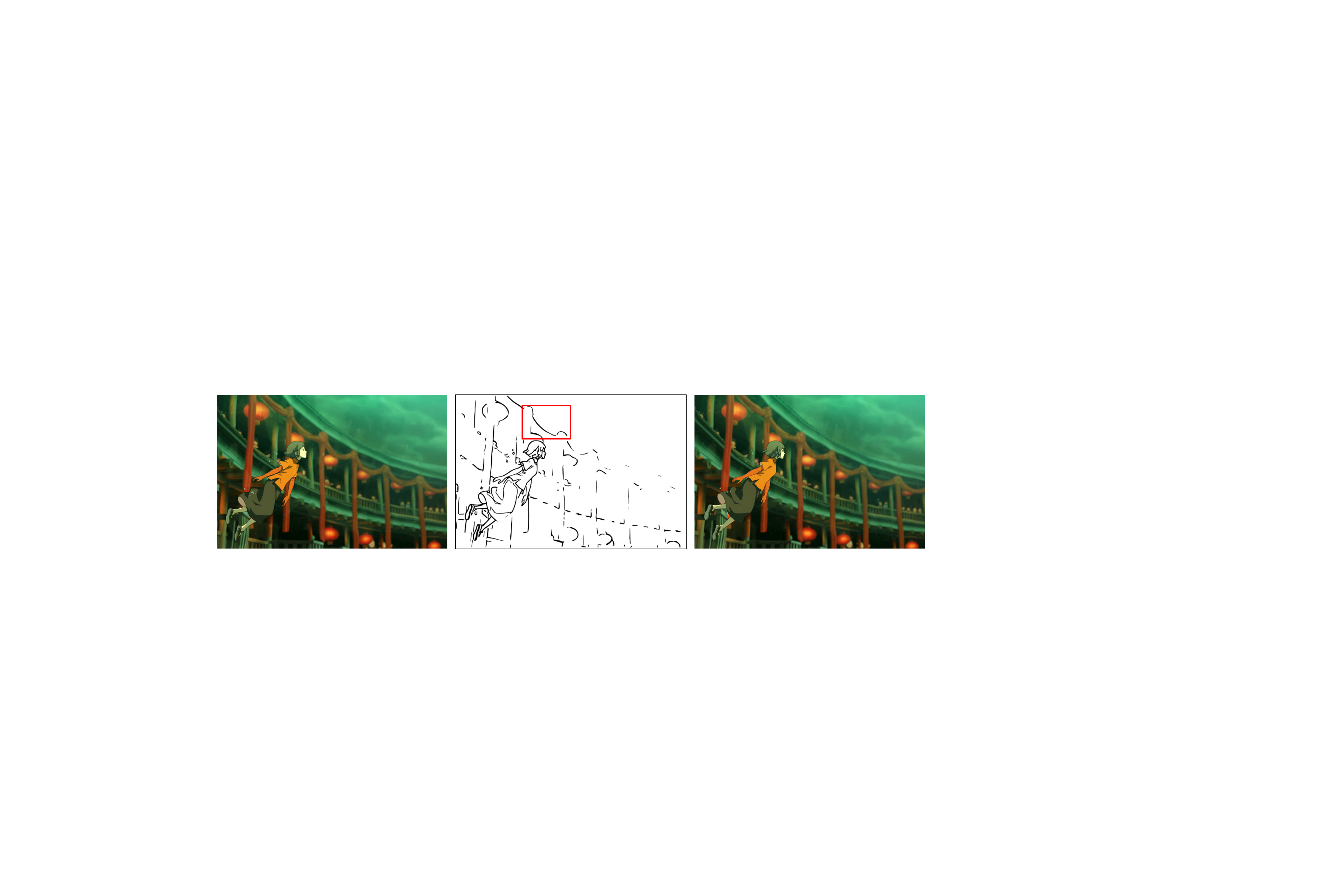}
	\begin{tabularx}{\linewidth}{p{0.01cm}YYY}
      & $I_0$ & $S_t$ & $I_t$ (ground truth)\\
    \end{tabularx}
	\includegraphics[width=\linewidth]{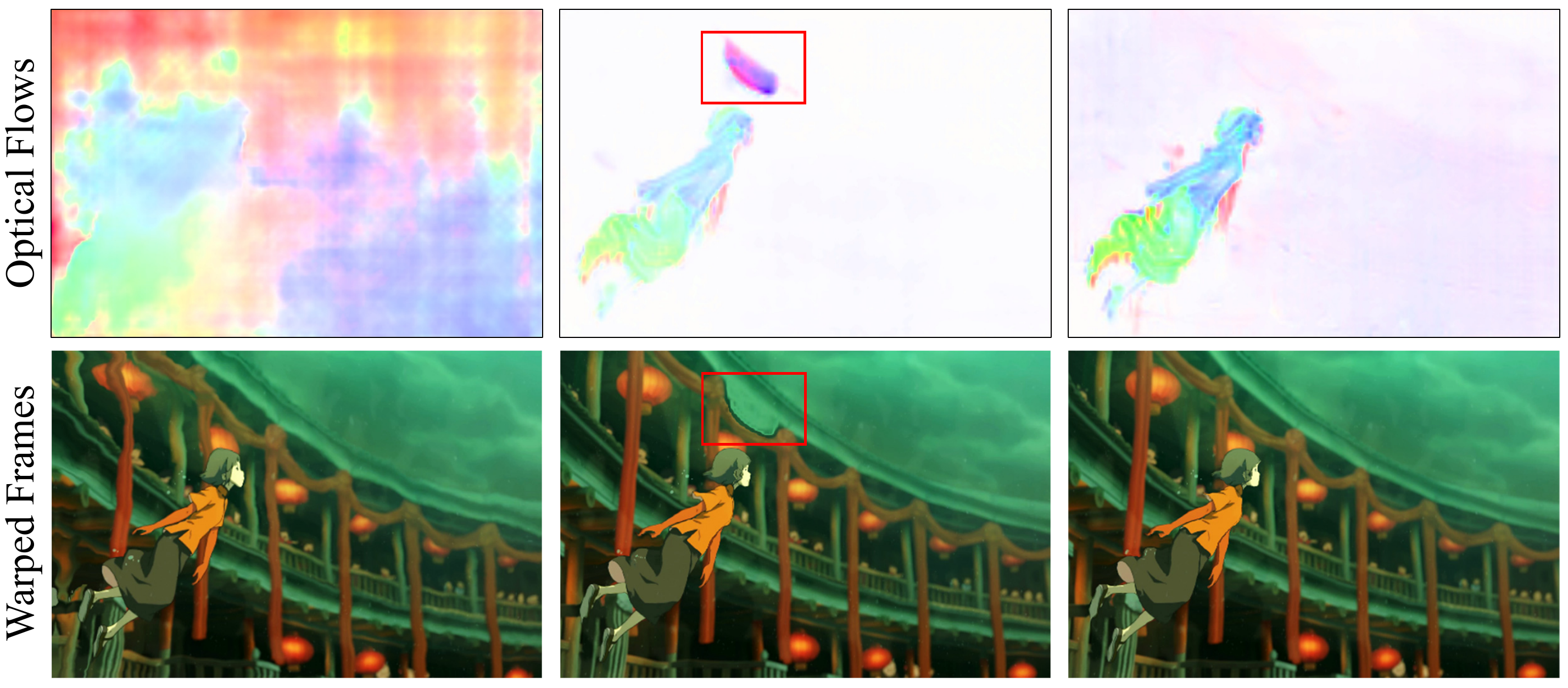}
	\begin{tabularx}{\linewidth}{p{0.01cm}YYY}
      & PWC-Net & PWC-Net  & Ours \\
      & \cite{sun2018pwc} & (fine-tune) & \\
    \end{tabularx}
	\caption[width=\textwidth]{Comparison of different flow estimation methods. Our method can handle movement in the presence of sparse sketches or empty regions. \textcopyright B\&T.}
	\label{fig:comp_flow}
\end{figure}

We first directly adopt PWC-Net~\cite{sun2018pwc} to estimate the cartoon-to-sketch correspondence. \change{In the second experiment, we fine-tune their model in our training set with cartoon sketch pairs as input and keep the network structure and the mechanism unchanged.} For our model, we remove the consistency checking and blending in the synthesis pipeline and only keep the cartoon-to-sketch correspondence network. The warping loss is utilized for both of these experiments. The quantitative results are shown in Table~\ref{tab:tab_est} and one example can be found in Figure~\ref{fig:comp_flow}. As the results show, our cartoon-to-sketch flow estimation method outperforms the common flow method by addressing challenging issues in texture-less cartoon frames and sketches with large empty regions.

\begin{table}
\centering
\small
\caption{Quantitative evaluation of cartoon-to-sketch correspondence estimation with different methods.}
\vspace{-0.3cm}
\label{tab:tab_est}
\setlength{\tabcolsep}{6pt}
\begin{tabular}{lccc}
	\toprule
	\multicolumn{1}{c}{\multirow{2}{*}{Model}} & \multicolumn{2}{c}{Sintel} & \multicolumn{1}{c}{2D Cartoon Clips}\\
	
	\cmidrule(lr){2-3} \cmidrule(lr){4-4}
	
    \multicolumn{1}{c}{} & EPE & $\ell_1$ Loss & $\ell_1$ Loss \\
	 
	 \cmidrule(lr){1-4}
	PWC-Net~\cite{sun2018pwc} & 22.55 & 0.0596 & 0.0336 \\
	PWC-Net (fine-tune)        & 10.93 & 0.0265 & 0.0112 \\
	\cmidrule(lr){1-4}
	our model  & \textbf{10.52} & \textbf{0.0247} & \textbf{0.0104}\\
	\bottomrule
\end{tabular}
\end{table}

\subsection{Comparison to Frame Interpolation Methods}

\begin{figure}
	\centering
    \small
	\includegraphics[width=\linewidth]{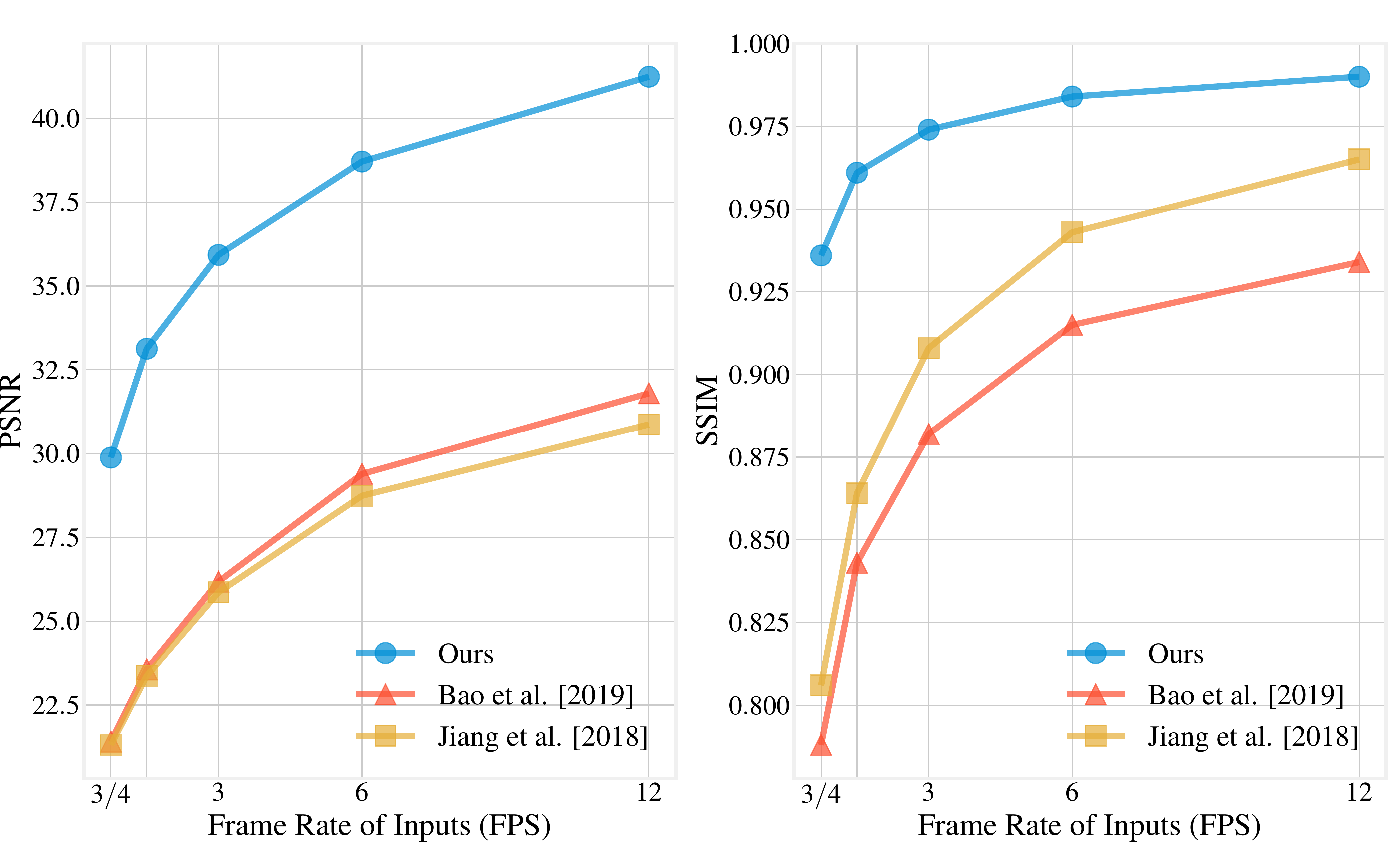}
	\caption{Quality comparison of our approach with other methods at different input frame rates. Lower frame rate means larger displacement between successive frames. Our method outperforms all other methods by a large margin by taking advantage of the guided sketch.}
	\label{fig:fps}
\end{figure}

\begin{figure}
    \scriptsize
	\centering
	\includegraphics[width=\linewidth]{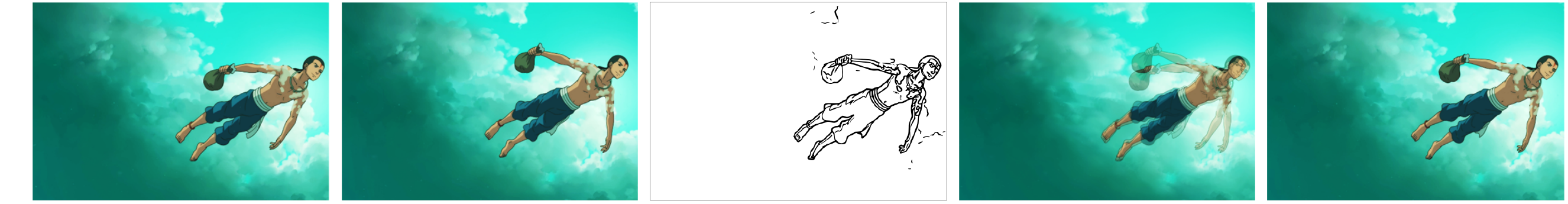}
	\begin{tabularx}{\linewidth}{p{0.02cm}YYYYY}
      & $I_0$ & $I_1$ & $S_{3/6}$ & \change{$I_{3/6}$ (Ave.)} & $I_{3/6}$ (GT) \\
    \end{tabularx}
	\includegraphics[width=\linewidth]{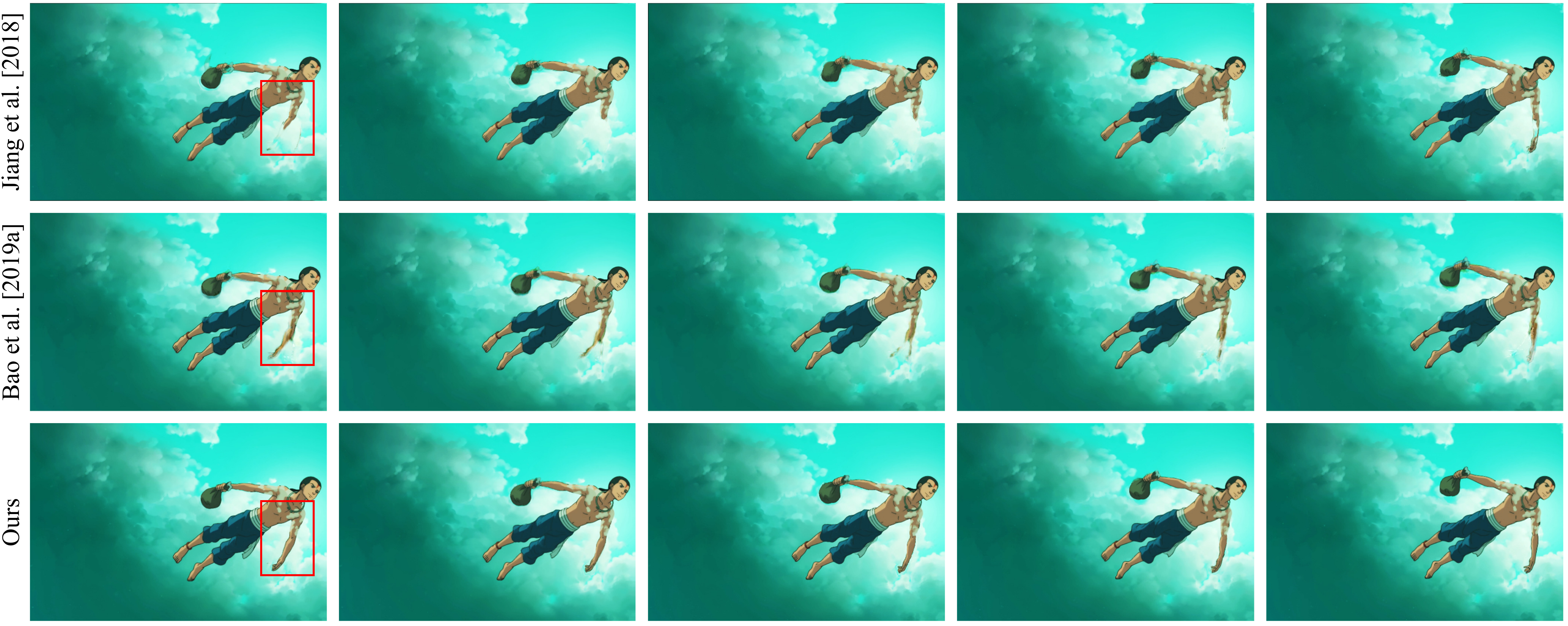}
	\begin{tabularx}{\linewidth}{p{0.02cm}YYYYY}
      & $\hat I_{1/6}$ & $\hat I_{2/6}$ & $\hat I_{3/6}$ & $\hat I_{4/6}$ & $\hat I_{5/6}$\\
    \end{tabularx}
	\caption{Example results comparing our method (fourth row) with other frame interpolation approaches: Jiang \etal~\cite{jiang2018super} (second row) and Bao \etal~\cite{bao2019depth} (third row). Compared to off-the-shelf methods, our method tends to better preserves the image content. \change{"Ave." represents average input frames and "GT" represents ground truth frame.} \textcopyright B\&T.}
	\label{fig:comp_int}
\end{figure}

\begin{figure*}
    \small
	\centering
	\includegraphics[width=\linewidth]{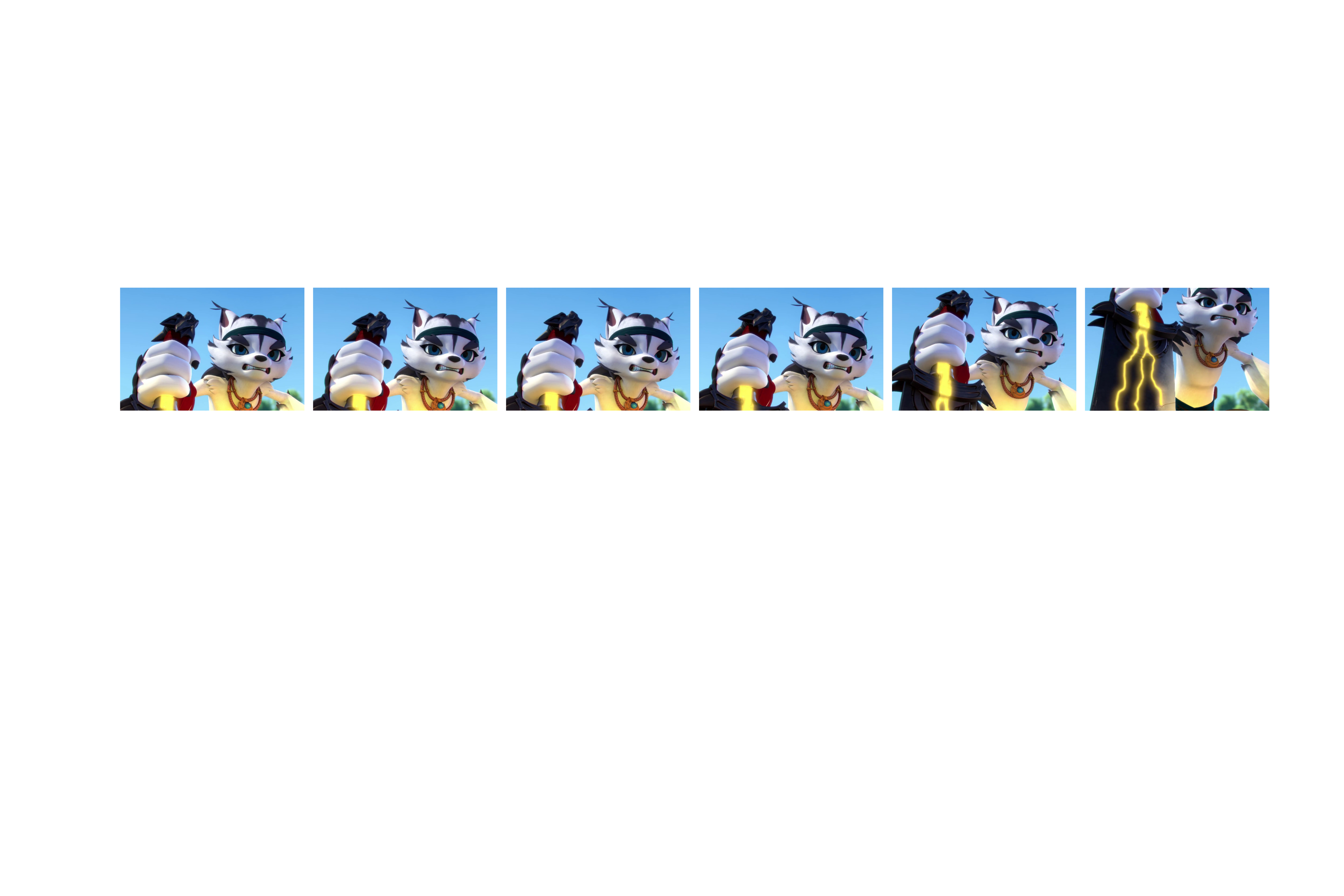}
	\begin{tabularx}{\linewidth}{YYYYYY}
        $I_0$ & $I_2$ & $I_4$ & $I_8$ & $I_{16}$ & $I_{32}$ \\
    \end{tabularx}
    \includegraphics[width=\linewidth]{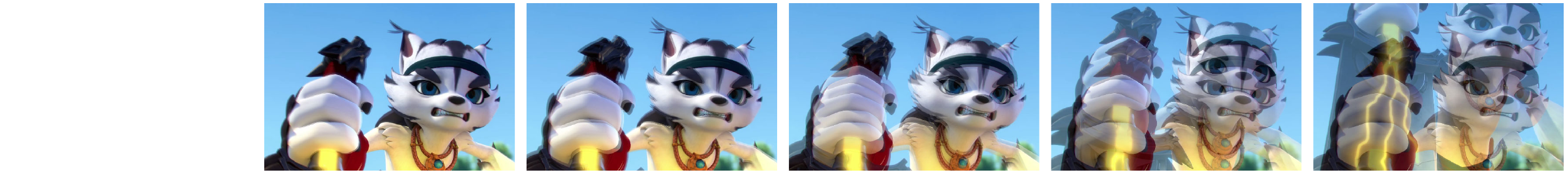}
	\begin{tabularx}{\linewidth}{YYYYYY}
         & \change{$\hat I_1$ (Ave.)} & \change{$\hat I_2$ (Ave.)} & \change{$\hat I_4$ (Ave.)} &   \change{$\hat I_8$ (Ave.)} & \change{$\hat I_{16}$ (Ave.)} \\
    \end{tabularx}
	\includegraphics[width=\linewidth]{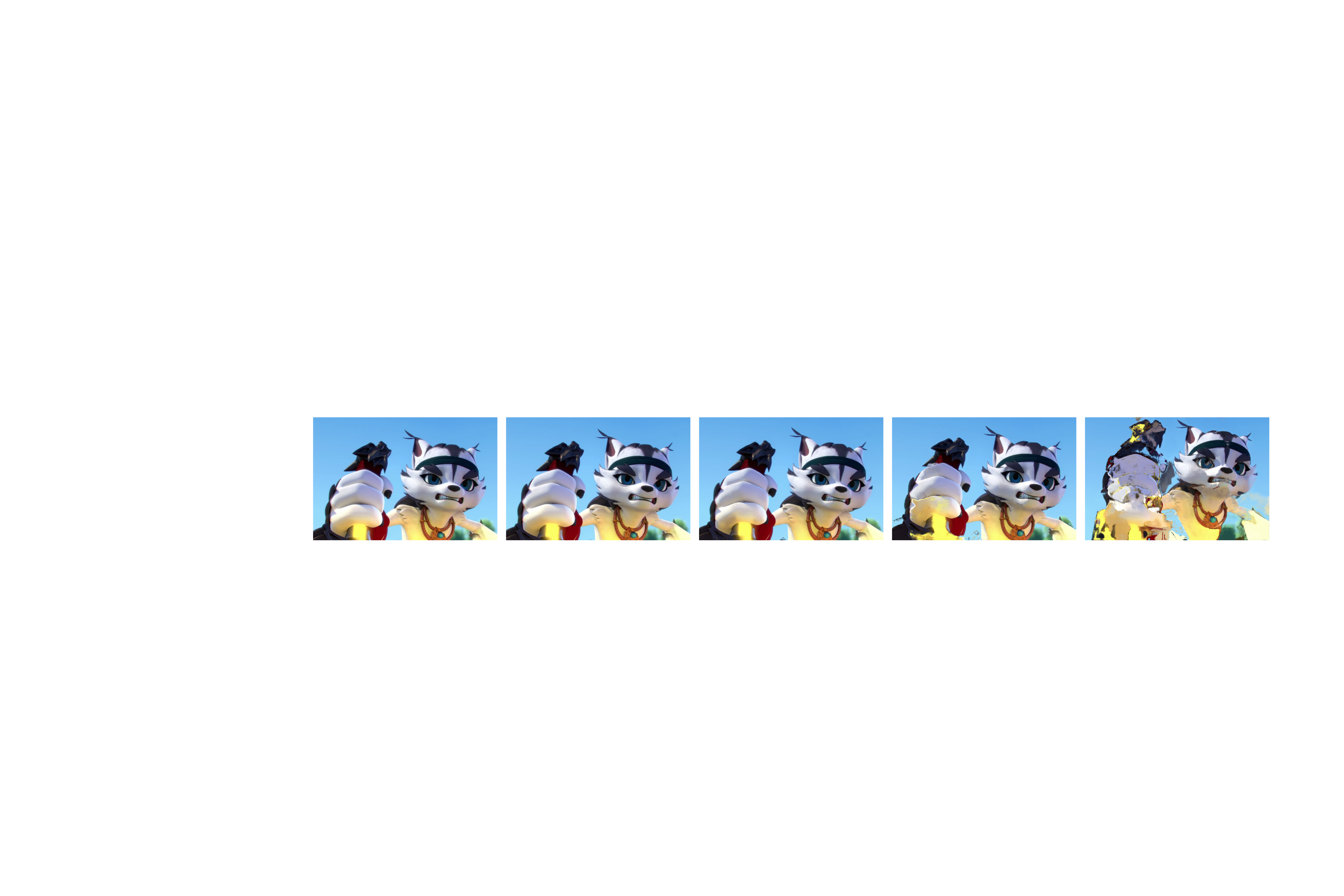}
	\begin{tabularx}{\linewidth}{YYYYYY}
         & $ \hat I_1$ & $\hat I_2$ & $\hat I_4$ & $\hat I_8$ & $\hat I_{16}$ \\
    \end{tabularx}
	\caption[width=\textwidth]{Example results with different input frame rates (deformation ranges). For each result $\hat I_k$ in the third row, $I_0$ and $I_{2k}$ from the first row are used as inputs. \change{We also show the averaged input frames for reference in the second row.} As the interval between input frames increases, artifacts begin to appear due to the larger motion. \textcopyright 2:10 AM Animation.}
	\label{fig:fps_example}
\end{figure*}

We next compare our method with some recent frame interpolation techniques that have source code available and are also able to interpolate arbitrary intermediate frames. These are the slow motion method by Jiang \etal~\cite{jiang2018super} (SloMo) and the depth-aware interpolation method by Bao \etal~\cite{bao2019depth} (DAIN). For a fairer comparison with these learning methods, we fine-tune their model on our training set. We use the {\em 3D Cartoon Clips} temporally downsampled to different frame rates as inputs for testing. 

The original frame rate for the clips is 24fps. We show results reconstructing the 24fps video using the input video temporally downsampled to different rates. \change{The other interpolation methods will take two adjacent frames of the downsampled video as input. For our method, in addition to the two input frames, we also utilize the sketch generated from the middle frame as additional guidance.} The quantitative results measuring PSNR and SSIM are shown in Figure~\ref{fig:fps}. Results show that our method outperforms all other methods by a large margin since it can take advantage of the sketch for guidance. The advantage is even more obvious when we interpolate frames with longer intervals. \change{We also show a qualitative comparison in Figure~\ref{fig:comp_int}, where our method takes $\{I_0, S_{3/6}, I_1\}$ as input to synthesize $I_{3/6}$ and interpolate four more frames $\{I_{1/6}, I_{2/6}, I_{4/6}, I_{5/6}\}$, and the other two methods take the two input frames $\{I_0, I_1\}$ to interpolate these five frames directly. We can see that both Jiang \etal~\cite{jiang2018super} and Bao \etal~\cite{bao2019depth} cannot always get satisfactory results due to the specific situations in cartoon frames, like large motion, texture-less style, unique contours. Our method takes advantage of the sketch as guidance, thus improving the the result by a large margin and also following the real-life workflow of cartoon animation. More importantly, the artists hope to control the inbetweening by drawing rather than using the deterministic result from interpolation. The sketch input provides the flexibility to control the interpolation result. We also shows an example of our results with different input frame rates in Figure~\ref{fig:fps_example} and it indicates that the lower the input frame rate, the lower the resulting quality.}

\begin{table}
\centering
\small
\change{\caption{Quantitative evaluation of frame interpolation only, providing ground truth to previous methods.}}
\vspace{-0.3cm}
\label{tab:tab_comp_int}
\setlength{\tabcolsep}{6pt}
\begin{tabular}{lcc}
	\toprule
    Method & PSNR & SSIM \\
	\cmidrule(lr){1-3}
	Averaged frames & 33.09 & 0.952 \\
	Jiang \etal~\cite{jiang2018super} & 29.85 & 0.955 \\
	Bao \etal~\cite{bao2019depth}     & 36.03 & \textbf{0.977} \\
	\cmidrule(lr){1-3}
	Ours  & \textbf{36.64} & 0.971 \\
	\bottomrule
\end{tabular}
\end{table}

\change{The additional sketch input of our method provides an advantage when we do the comparisons to earlier work,  showing the need for such a sketch in order to achieve the artist's intent. However, this is an advantageous comparison since the previous methods do not use a sketch and are just focused on interpolation. Thus, we provide another experiment to test just the interpolation ability of our method in a disadvantageous condition to our approach. We use the {\em 3D Cartoon Clips} in Section 5.1 as the test data for this experiment. For these clips, we take every 7 consecutive frames  $I_{0/6}$ to $I_{6/6}$ as a sample. We use $\{I_0, S_{3/6}, I_1\}$ to interpolate $\{I_{1/6}, I_{2/6}, I_{4/6}, I_{5/6}\}$ for our method and use $\{I_0, I_{3/6}, I_1\}$ to interpolate $\{I_{1/6}, I_{2/6}, I_{4/6}, I_{5/6}\}$ for the two interpolation methods. We also compared the averaged input frames. The result is shown in Table 4. We can see that even if this experiment is disadvantageous for our method since we only have $S_{3/6}$ while the others have ground truth $I_{3/6}$ for interpolation, our method can still produce competitive results in terms of interpolation ability.}

\subsection{Comparison to Image Synthesis Methods}

\begin{figure*}
	\centering
    \small
	\includegraphics[width=\linewidth]{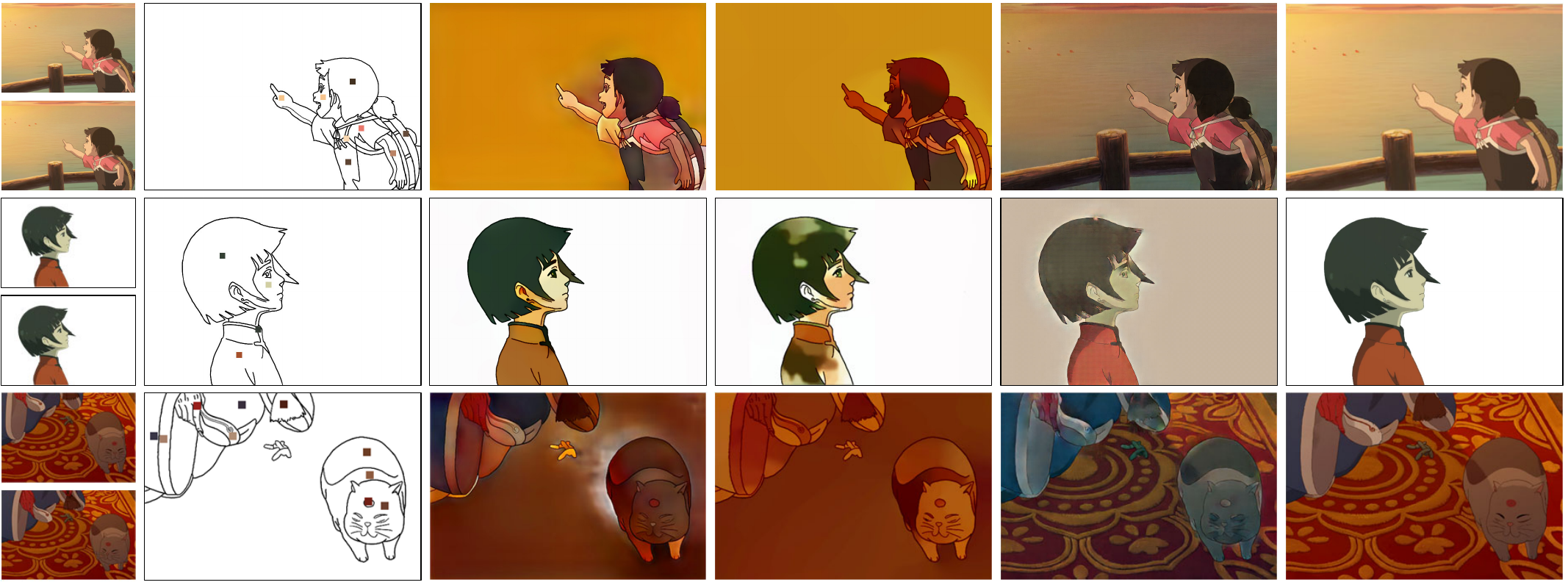}
	\begin{tabular}{x{0.07\linewidth}x{0.16\linewidth}x{0.16\linewidth}x{0.16\linewidth}x{0.16\linewidth}x{0.16\linewidth}}
		\change{$I_0$ $I_1$} & \change{$S_t$ and hints (only for Zhang \etal~\cite{zhang2018two})} & \change{Zhang \etal~\cite{zhang2018two} by using hints} & \change{Zhang \etal~\cite{zhang2018two}  by using references} & \change{Wang \etal~\cite{wang2018high}} & \change{Ours} \\
	\end{tabular}
	\caption{Example results comparing image synthesis methods with our approach. \textcopyright B\&T.}
	\label{fig:comp_gan}
\end{figure*}

We compare our method with recent image generation or synthesis techniques. More specifically, we compare it with pix2pixHD~\cite{wang2018high}, a state-of-the-art conditional generative adversarial networks and a sketch colorization method~\cite{zhang2018two} which also targets cartoon images and utilizes two-stage training strategy. For the pix2pixHD, we use the sketch image as the input and two cartoon frames as the image translation condition. The model is trained using our training set until convergence. For the sketch colorization method, we directly use their pre-trained model for inference as their method is trained on a large-scale cartoon dataset. \change{We show two results of their method. One is to use the manually selected color hints to colorize the sketch. Another is to use the cartoon frame as the reference to colorize the sketch.}

\change{As shown in Figure~\ref{fig:comp_gan}, the sketch colorization method is unable to handle the complex color styles and tends to use a relatively smooth color within a region even using many color hints as additional inputs. Moreover, their method does not utilize the temporal information of the video, thus produce a relatively low-quality result. Due to the nature of colorization, it faithfully follows edges from the input sketch but cannot recover any structure details that are missing the sketch. Our method has the ability to address such variations, such as the shadow on the clothes in the second example.} On the other hand, pix2pixHD suffers from bleeding artifacts in the regions with large motion. Furthermore, it severely overfits the cartoon style in the training set: when we test on a movie beyond the training set, it suffers from color shifting.

\subsection{Generalization Ability and Flexibility}

\begin{figure*}
    \small
	\centering
\includegraphics[width=\linewidth]{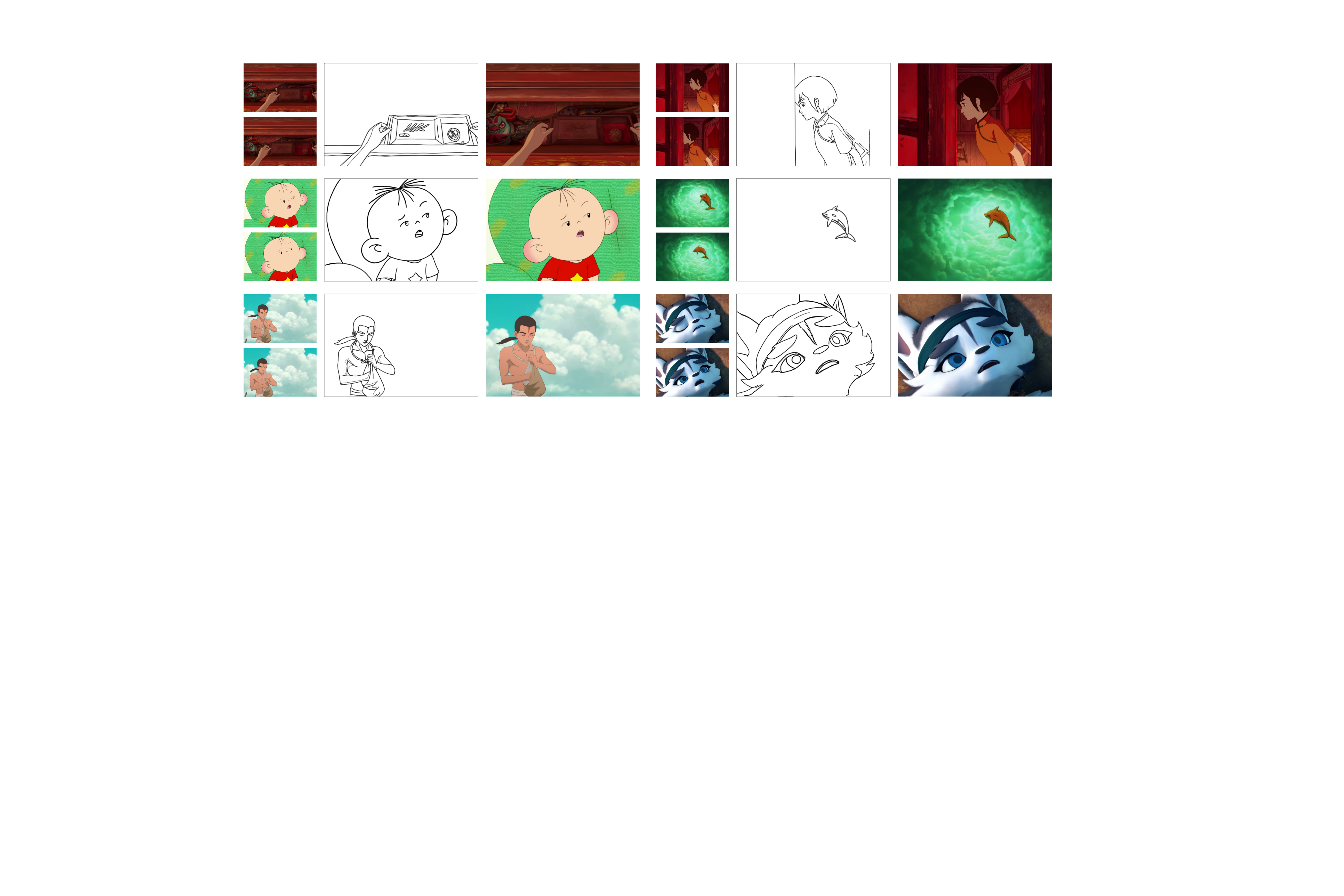}
	\begin{tabular}{x{0.08\linewidth}x{0.18\linewidth}x{0.18\linewidth}x{0.08\linewidth}x{0.18\linewidth}x{0.18\linewidth}}
		$\{I_0, I_1\}$ & $S_t$ & $\hat I_t$ & $\{I_0, I_1\}$ & $S_t$ & $\hat I_t$ \\
	\end{tabular}
	\caption[width=\textwidth]{Results from hand-drawn sketches with different cartoon and drawing styles. \textcopyright B\&T, 2:10 AM Animation, SAFS.}
	\label{fig:res_more}
\end{figure*}

\begin{figure*}
    \small
	\centering
	\includegraphics[width=\linewidth]{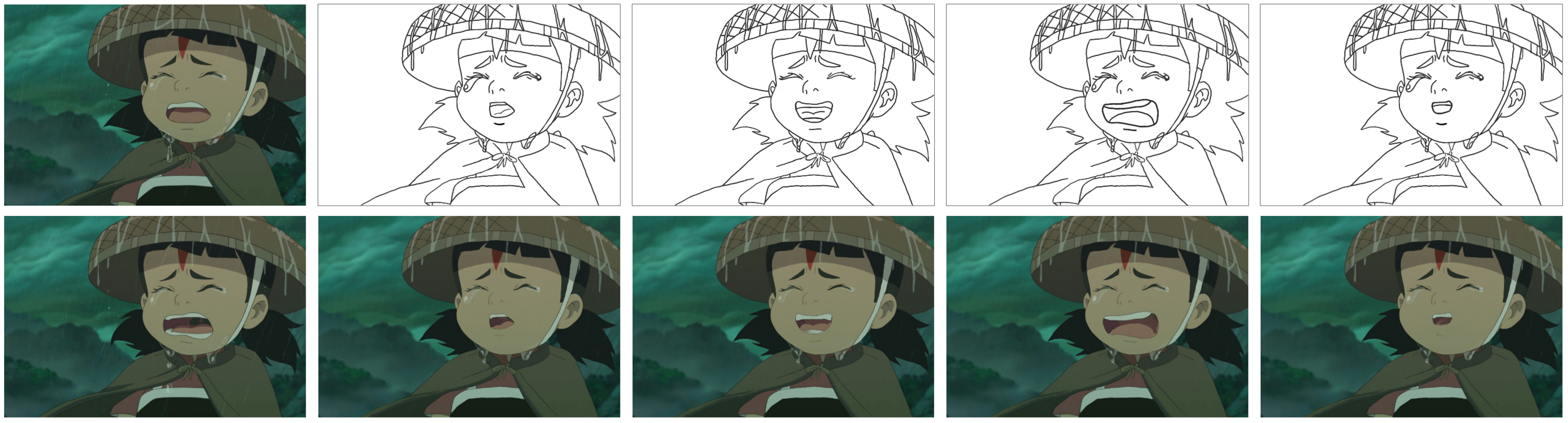}
	\caption[width=\textwidth]{Our system can synthesize different middle frames by drawing different guided sketches. In this example, the proposed method takes two input frames (first column) and synthesizes the cartoon frames (second row) guided by the corresponding user's sketches (first row). \textcopyright B\&T.}
	\label{fig:res_multiskt}
\end{figure*}


\begin{figure}
    \small
	\centering
	\includegraphics[width=\linewidth]{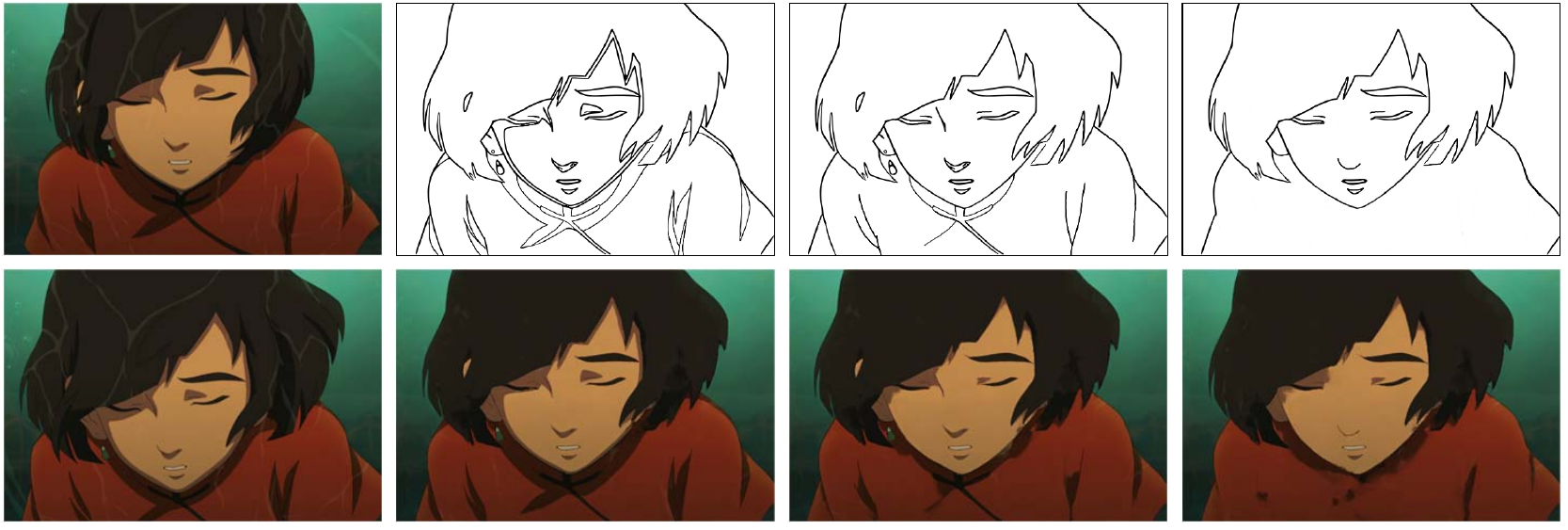}
	\begin{tabularx}{\linewidth}{YYYY}
        $\{I_0, I_1\}$ & First level & Second level & Third level \\
    \end{tabularx}
	\change{\caption[width=\textwidth]{Results using sketches with different levels of detail. In this example, the proposed method takes two input frames (first column) and synthesizes the cartoon frames (second row) guided by the corresponding sketches (first row). \textcopyright B\&T.}}
	\label{fig:res_multilevel}
\end{figure}

We try to maximize the generalization of our method by constructing a dataset containing diverse scenes and large motions. Furthermore, we attempt to synthesize simplified sketches as realistically as possible. Though ultimately we trained the network solely using the training samples from only one 2D cartoon movie and one 3D cartoon animations, our method can still perform well on frames in movies with different styles, even a 3D cartoon that does not have obvious contours and 2D animations as shown in Figure~\ref{fig:res_more}. Our training set only contains one single sketch style synthesized by our algorithm. However, it still has the capability to generalize to rough sketches after simplification (Figure~\ref{fig:teaserfigure}) or hand-drawn sketches (Figure~\ref{fig:res_more}, ~\ref{fig:res_multiskt}). Furthermore, our method has the flexibility to generate different results by providing different guided sketches as shown in Figure~\ref{fig:res_multiskt}. Users can choose to automatically interpolate the whole video by just drawing one middle sketch if the motion is relatively simple, or drawing more sketches to synthesize frame by frame if the motion is complex and can not be interpolated. The user can also use a combination of the two approaches. \change{In Figure 15, we use an example to show how the level of detail affects the output of our result. To do so, we progressively remove some lines from the first level of the real sketch and show their results in the second row. Note that our method can still produce reasonable results when omitting some lines (e.g. remove the shadow of the hair on the face or change the nose according to the lines). But artifacts may appear if some important lines indicating the motion are omitted. The model can produce a better result if more details are provided in the sketch. Moreover, due to the fully convolutional networks we use, our method can address video frames with different resolutions without a drop in performance. One example can be found in Figure~\ref{fig:res_more}, whose original frames come from different movies at different resolutions (e.g., 480p, 720p, and 1080p).}

The capabilities of our method can be summarized as follows:
\begin{enumerate}
\item It generalizes to different cartoon styles;
\item \change{It generalizes to sketches with some variations};
\item It supports the generation of different video results from the same input by drawing different sketches;
\item It supports drawing and synthesizing one (or more) sketch and interpolating the remaining frames;
\end{enumerate}

\section{Limitations and Conclusion}

\begin{figure}
	\centering
    \small
	\includegraphics[width=\linewidth]{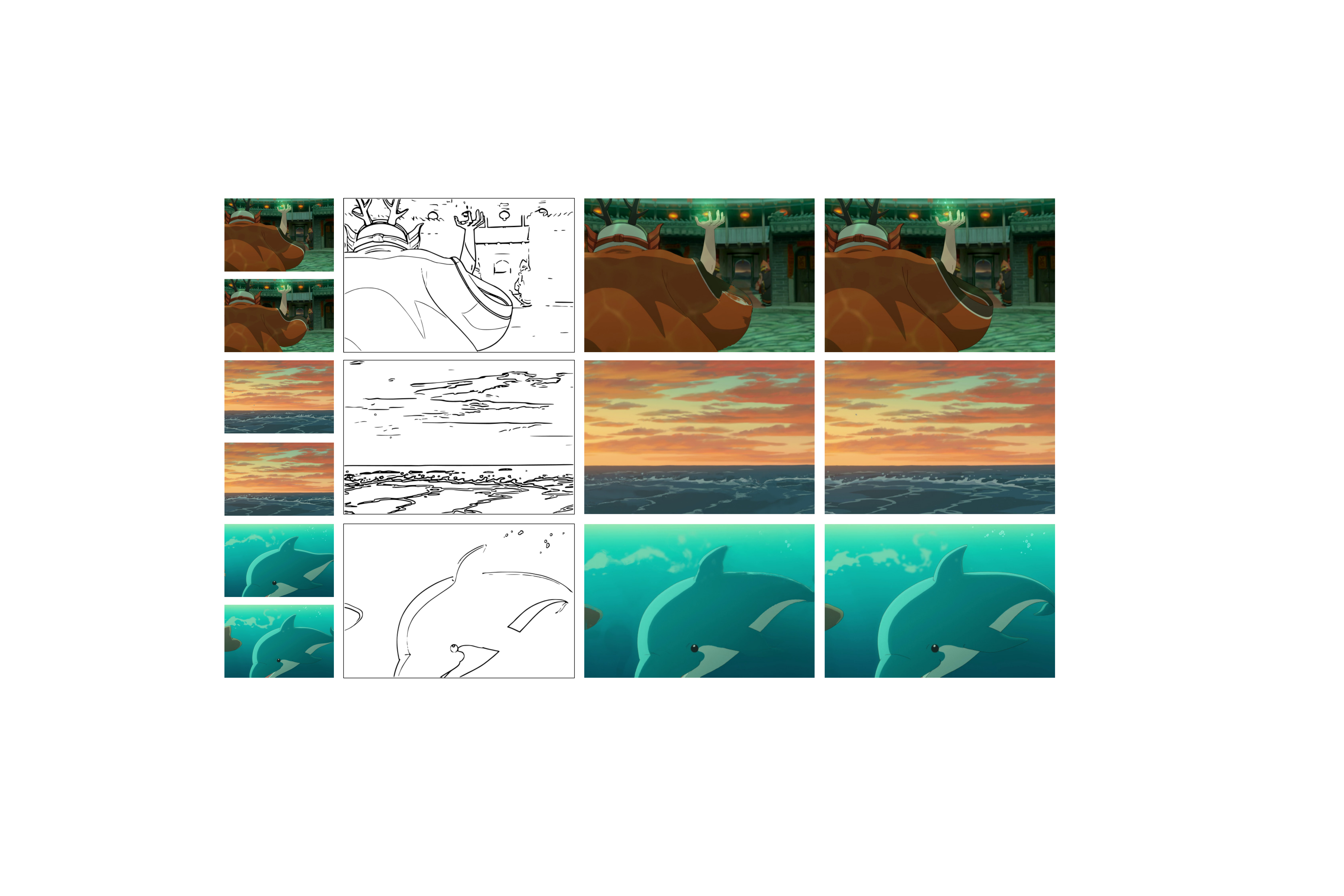}
	\begin{tabular}{x{0.12\linewidth}x{0.24\linewidth}x{0.24\linewidth}x{0.24\linewidth}}
	$I_0$ $I_1$ & $S_t$ & $I_t$ (ours) & $I_t$ (ground truth)\\
    \end{tabular}
	\caption{Examples where our approach did not yield satisfactory results, including pixels being occluded in both two input frames, unclear semantic correspondence, and artifacts due to the incomplete sketch for indicating motion. \textcopyright B\&T.}
	\label{fig:limitations}
\end{figure}

In conclusion, we present a novel framework that synthesizes cartoon videos by using the color information from two input frames while following the animated motion guided by a sketch. Our approach first estimates the dense cross-domain correspondence between a sketch and video frames by transforming the cartoon and sketch into feature representations in the same domain, followed by applying a blending module for occlusion handling considering flow consistency. Then, the inputs and the synthetic frame equipped with established correspondence is fed into an arbitrary-time interpolation pipeline to generate and refine more inbetween frames. Finally, a video temporal processing approach is used to further improve the result. We perform several experiments to verify each component of our system, show side-by-side comparisons with related methods, and demonstrate the generalization ability and flexibility of our system. Our results show that our system generalizes well to different scenes and produce high-quality results. 

However, there are some cases that our method cannot handle perfectly. First, our method is based on warping and blending. If the pixels in the middle frame are occluded in both two input frames, artifacts will appear as shown in the first example of Figure~\ref{fig:limitations}. Second, some scenes in a 2D cartoon without accurate semantic correspondence, such as the waves in Figure~\ref{fig:limitations} which can appear and disappear suddenly with different shapes, also cannot be addressed by our method. Third, when the contours that are vital to indicate motion are missing in the sketch image, it is hard for our method to infer that information from the two input frames accurately, e.g., the fins of the dolphin in the third example of Figure~\ref{fig:limitations}. To address this limitation, our method allows the user to interactively drawing more strokes in the sketch. It would be worthwhile to explore how to solve these artifacts automatically.

\section*{Acknowledgments}
This work was supported in part by the Hong Kong Research Grants Council (RGC) Early Career Scheme under Grant CityU 21209119, the CityU of Hong Kong under APRC Grant 9610488, and at HKUST by grants FSGRF16EG08 and DAG06/07.EG07.

\newpage
\bibliographystyle{IEEEtran}
\bibliography{main}

%

\begin{IEEEbiography}
[{\includegraphics[width=1in,height=1.25in,clip,keepaspectratio]{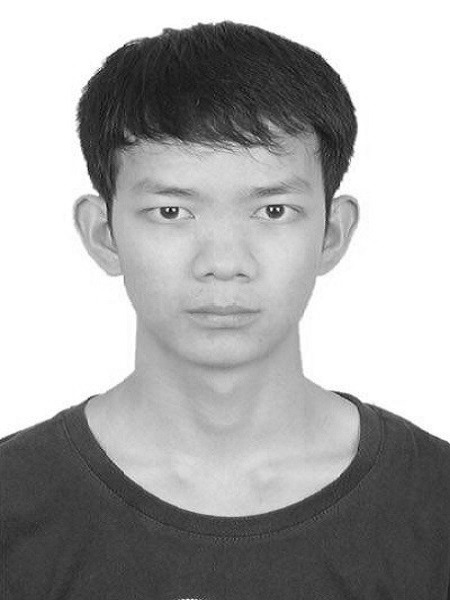}}]{Xiaoyu Li}
received a Bachelor of Engineering degree in Electronic Information Engineering from Huazhong University of Science and Technology, in 2017. He is currently pursuing a Ph.D. degree with the Electronic and Computer Engineering, Hong Kong University of Science and Technology. His research interests include computer vision and deep learning with an emphasis on computational photography.
\end{IEEEbiography}

\begin{IEEEbiography}
[{\includegraphics[width=1in,height=1.25in,clip,keepaspectratio]{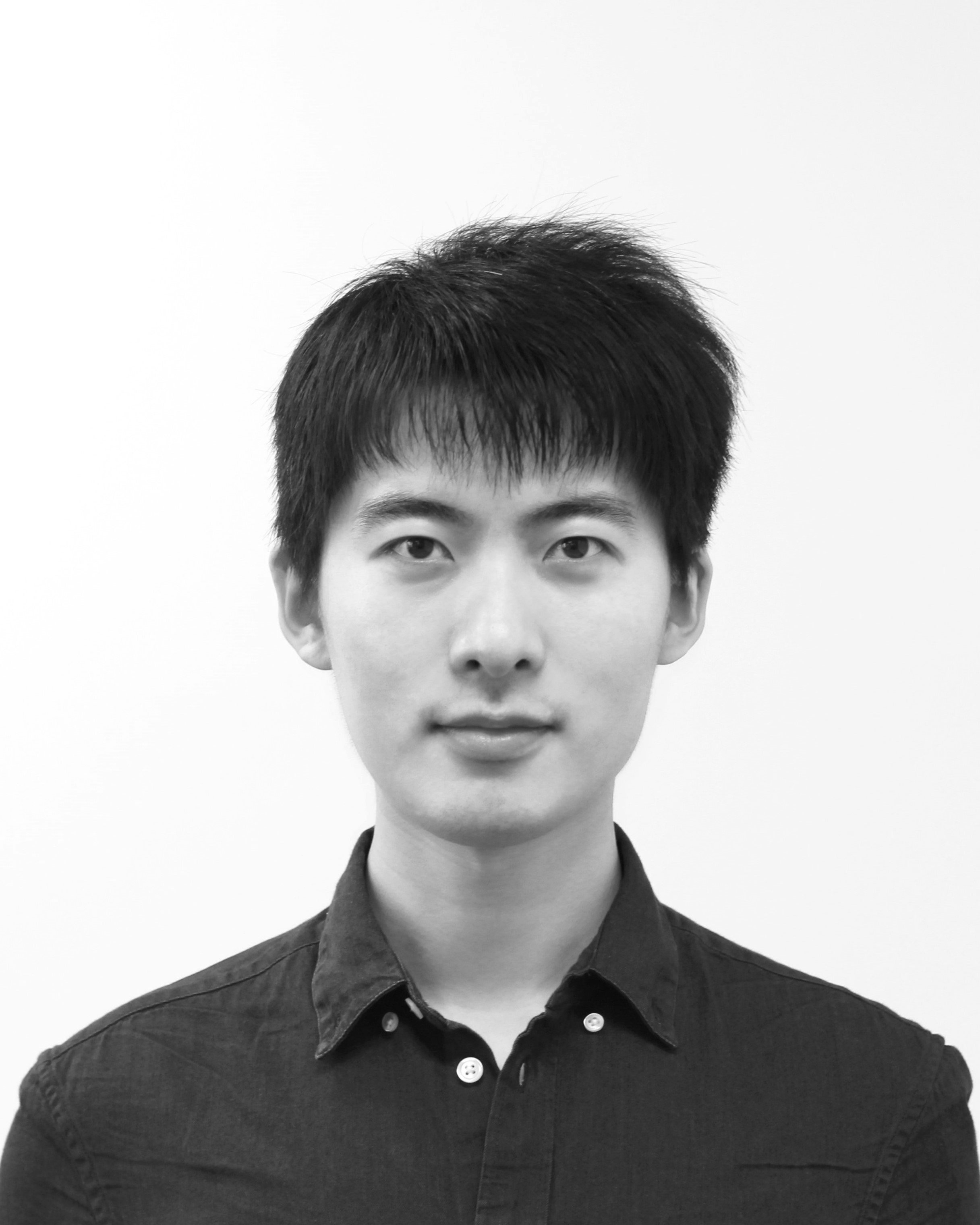}}]
{Bo Zhang}
received his Ph.D. degree with the Department of Electronic and Computer Engineering at the Hong Kong University of Science and Technology at 2019. Prior to that, he received a Bachelor of Engineering degree at Zhejiang University. Now he is a researcher at visual computing group of Microsoft research asia. His research interests involve low-level computer vision, image synthesis, computational photography and imaging system.
\end{IEEEbiography}

\begin{IEEEbiography}
[{\includegraphics[width=1in,height=1.25in,clip,keepaspectratio]{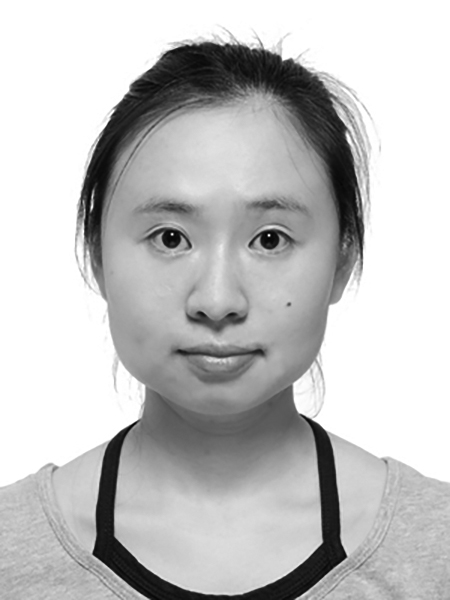}}]{Jing Liao}
is an Assistant Professor with the Department of Computer Science, City University of Hong Kong (CityU) since Sep 2018. Prior to that, she was a Researcher at Visual Computing Group, Microsoft Research Asia from 2015 to 2018. She received the B.Eng. degree from Huazhong University of Science and Technology and dual Ph.D. degrees from Zhejiang University and Hong Kong UST. Her primary research interests fall in the fields of Computer Graphics, Computer Vision, Image/Video Processing, Digital Art and Computational Photography.
\end{IEEEbiography}

\begin{IEEEbiography}
[{\includegraphics[width=1in,height=1.25in,clip,keepaspectratio]{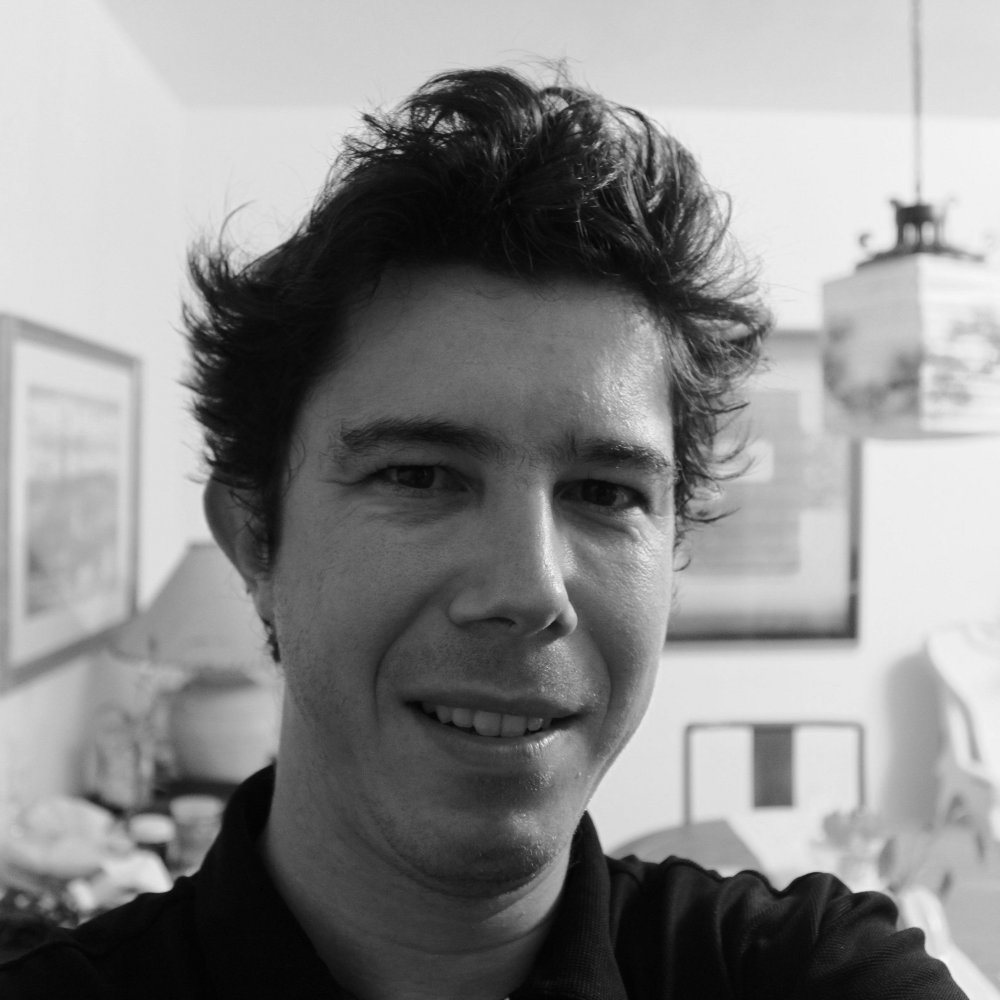}}]{Pedro V. Sander}
received a Bachelor of Science in Computer Science from Stony Brook University, and Master of Science and Doctor of Philosophy degrees from Harvard University. He was a senior member of the Application Research Group of ATI Research, where he conducted real-time rendering and general-purpose computation research with latest generation and upcoming graphics hardware. Currently, he is a Professor in the Department of Computer Science and Engineering at the Hong Kong University of Science and Technology. His research interests lie mostly in real-time rendering, graphics hardware, geometry processing, and imaging. 
\end{IEEEbiography}




\end{document}